\documentclass[journal]{IEEEtran}
\usepackage{comment, graphicx, amsmath, algcompatible, algorithm, epstopdf,epsfig,amssymb,url,multirow, graphicx, enumerate,float,diagbox,makecell,ragged2e,color,array, cancel}
\DeclareGraphicsExtensions{.png,.pdf}
\graphicspath{{pic/}}

\usepackage{subfigure}
\usepackage{caption}
\usepackage{cite}
\begin{document}

\title{Image Copy-Move Forgery Detection via Deep PatchMatch and Pairwise Ranking Learning}
\author{
{Yuanman Li,~\IEEEmembership{Senior Member,~IEEE,} 
Yingjie He,~\IEEEmembership{Student Member,~IEEE,} 
Changsheng Chen,~\IEEEmembership{Senior Member,~IEEE,}
Li Dong, 
Bin Li,~\IEEEmembership{Senior Member,~IEEE,}
Jiantao Zhou,~\IEEEmembership{Senior Member,~IEEE,} and 
Xia Li,~\IEEEmembership{Member,~IEEE}
}
\thanks{Y. Li, Y. He, C. Cheng, B. Li and X. Li are with Guangdong Key Laboratory of Intelligent Information Processing, College of Electronics and Information Engineering, Shenzhen University, Shenzhen, China. 

L. Dong is with the Department of Computer Science, Ningbo University, Ningbo, China. (email: dongli@nbu.edu.cn).

J. Zhou is with the Department of Computer and Information Science, University of Macau, Macau, China. (e-mail: jtzhou@um.edu.mo).
}}
\maketitle

\begin{abstract}
Recent advances in deep learning algorithms have shown impressive progress in image copy-move forgery detection (CMFD). However, these algorithms lack generalizability in practical scenarios where the copied regions are not present in the training images, or the cloned regions are part of the background. Additionally, these algorithms utilize convolution operations to distinguish source and target regions, leading to unsatisfactory results when the target regions blend well with the background. To address these limitations, this study proposes a novel end-to-end CMFD framework that integrates the strengths of conventional and deep learning methods. Specifically, the study develops a deep cross-scale PatchMatch (PM) method that is customized for CMFD to locate copy-move regions. Unlike existing deep models, our approach utilizes features extracted from high-resolution scales to seek explicit and reliable point-to-point matching between source and target regions. Furthermore, we propose a novel pairwise rank learning framework to separate source and target regions. By leveraging the strong prior of point-to-point matches, the framework can identify subtle differences and effectively discriminate between source and target regions, even when the target regions blend well with the background. Our framework is fully differentiable and can be trained end-to-end. Comprehensive experimental results highlight the remarkable generalizability of our scheme across various copy-move scenarios, significantly outperforming existing methods.
\end{abstract}
\IEEEpeerreviewmaketitle
\section{Introduction}
	\label{sec:intro}
The proliferation of digital image editing tools has made image forgeries increasingly prevalent in our daily lives. These manipulated images can be maliciously exploited in activities such as internet rumors, insurance fraud, fake news dissemination, and even academic dishonesty, posing significant security concerns to society. Copy-move forgery, a prevalent form of manipulation, involves duplicating and relocating objects within an image to alter its content. Detecting such forgeries is challenging due to the similarity in statistical characteristics between the forged and untouched regions, including noise distribution, brightness, and photometric attributes.

Image copy-move forgery detection (CMFD) is a significant focus in multimedia security. It has witnessed substantial research efforts in recent years, yielding diverse proposed approaches. Traditional CMFD techniques utilize hand-crafted features to identify copy-move correspondences, including block-based methods\cite{bashar_exploring_2010,ryu_rotation_2013,lee_detection_2015,pun_two-stage_2018} and keypoint-based methods\cite{amerini_sift-based_2011,manu_detection_2016,li_fast_2019}. While these algorithms offer credible results by explicitly linking forged traces through block or keypoint matching, their reliance on manually crafted features tailored for generic vision tasks limits their efficacy against complex copy-move forgeries and susceptibility to post-processing. Moreover, these conventional methods solely detect correspondences and fail to distinguish between source and target regions.

Motivated by the potent representational capabilities of deep features, recent years have witnessed a surge in the study of deep CMFD frameworks. Wu \textit{et al.} \cite{ferrari_busternet_2018} introduced the pioneering end-to-end CMFD framework with source/target separation. Unlike traditional methods, \cite{ferrari_busternet_2018} harnesses convolutional neural networks (CNNs) to adaptively learn features from a CMFD dataset, obviating the need for manual design. This innovation has spurred the development of subsequent deep CMFD models, including \cite{liu2021two,chen_serial_2020, islam_doa-gan_2020}, all of which leverage the insights from \cite{ferrari_busternet_2018}. Armed with these representative features, these deep CMFD approaches exhibit heightened real-world effectiveness and enhanced resilience against post-processing.
However, despite these merits, they still exhibit certain inherent limitations, as outlined below:
\begin{itemize}
    \item Most existing deep Copy-Move Forgery Detection (CMFD) methods identify potential copy-move regions by generating attention maps using high-level features from deep CNNs. Nevertheless, as will be verified in our experiments, these features often become overfitted to objects in the training images,  leading to a significant decrease of models' generalizability. Consequently, these approaches are less effective in cases where copy-moved elements are missing from the training data. Moreover, they may fail entirely when the copy-move manipulation occurs in the background. 
    \item In contrast to traditional methods, deep CMFD techniques struggle to establish explicit point-to-point matching for copy-move correspondences. Consequently, the interpretability and reliability of their detection outcomes are compromised.
    \item They simply apply convolution operations on the entire input image for source/target separation, which cannot effectively utilize the strong point-to-point relationship between source and target regions. Consequently, they often fail to separate source/target regions when the target regions blend well with the background.
\end{itemize}

In response to the aforementioned challenges, we present a novel end-to-end framework for CMFD, named Deep PM and Pairwise Ranking Learning (D2PRL). D2PRL combines strengths from both deep models and conventional techniques. First, D2PRL focuses on capturing point-to-point matching between source and target regions in high-resolution scales, enhancing its generalizability across various types of copy-move contents. Second, D2PRL is fully differentiable, enabling features to be learned through an end-to-end training process. Finally, D2PRL can effectively discriminate source/target regions with the assistance of strong prior knowledge of point-to-point matches. 
The primary contributions of our work are summarized below:
\begin{itemize}
\item We introduce an innovative end-to-end deep CMFD framework featuring source/target separation, harnessing the strengths of conventional and modern deep CMFD models. Our approach significantly outperforms existing algorithms and demonstrates very robust generalizability to various copy-move content, encompassing objects and backgrounds.

\item We devise a fine-grained similarity localization method based on deep cross-scale PM customized for CMFD. Additionally, several techniques, such as cross-scale matching and multi-scale dense fitting error estimation are developed for seeking reliable point-to-point matching between source and target regions in high resolution scales. 


\item By leveraging the strong prior of point-to-point matches, we propose converting the original problem of discriminating source and target regions into a pairwise ranking problem. This approach compels the network to uncover subtle clues for distinguishing between the source and target regions, even when the target regions blend well with the background.
\end{itemize}

The rest of this paper is structured as follows: Section \ref{sec:relatedworks} provides a brief review of related works, while Section \ref{sec:proposed} elaborates on our CMFD framework. Section \ref{sec:Experiment Result} presents extensive experimental results and ablation studies, and Section \ref{sec:Conculsion} concludes the paper.


\section{Related Works} \label{sec:relatedworks}
In this section, we give brief reviews of the related works, including existing CMFD methods and the PM algorithm.

\subsection{Image copy-move forgery detection}
Previous conventional CMFD algorithms adopt manually designed features to identify the near duplicated regions. They primarily follow a common pipeline, i.e., 1) feature extraction for each location or keypoint, 2) feature matching to reveal copy-moved pixels,  and 3) post-processing to refine the detected regions. These approaches can be categorized into block-based algorithms and keypoint-based algorithms, based on their feature extraction and matching methods. Block-based methods first split the image into overlapped regions, and then apply the matching algorithm to identify near-duplicated regions. The designed features are required to be robust against common transformations, such as rotation and scaling. In previous years, a number of feature extraction schemes have been investigated in block-based algorithms, including Discrete Cosine Transform (DCT) coefficients \cite{lee_detection_2015}, discrete wavelet transform (DWT) \cite{bashar_exploring_2010}, Radial Harmonic Fourier moments (RHFMs) \cite{zhong_radon_2016}, Analytic Fourier-Mellin transform (AFMT) \cite{pun_two-stage_2018} and Zernike Moments (ZM) \cite{ryu_rotation_2013}. For instance, Ryu \textit{et al.} \cite{ryu_rotation_2013} conducted feature matching using locality-sensitive hashing (LSH) and utilized rotationally invariant ZM features to detect rotated copy-move areas. Emam \textit{et al.} \cite{emam_pcet_2016} extracted features of circular image blocks through the Polar Complex Exponential Transform (PCET) and employed Approximate Nearest Neighbor (ANN) to complete the circular block matching process, thereby identifying potential copy-move regions.
Different from block-based methods, keypoint-based algorithms extract features only for some points of interest sparsely located in the image and perform keypoint matching to reveal suspicious point pairs. Speeded-Up Robust Feature (SURF) \cite{manu_detection_2016} and Scale Invariant Feature Transform (SIFT) \cite{amerini_sift-based_2011,li_fast_2019} are the mostly studied feature extraction algorithms for keypoint-based approaches.  For example, Pan and Lyu \cite{pan_region_2010} were pioneers in using SIFT feature-based keypoint matching for copy-move forgery detection. Their proposed method exhibited strong robustness against geometric transformations.
Shivakumar \textit{et al.} \cite{shivakumar2011detection} improved the efficiency of detection algorithms by completing keypoint matching through SURF and KD-tree. 
Pun \textit{et al.} \cite{pun_image_2015} integrated keypoint-based and block-based detection algorithms. They first segmented the image into non-overlapping and irregular blocks using an adaptive over-segmentation algorithm, and then extracted keypoint features from the image blocks and located potential tampered areas through a block matching algorithm. Wang \textit{et al.} \cite{10007894} alleviated the issue of difficulty in extracting keypoints in homogeneous regions by removing the contrast threshold and increasing the image resolution, and introduced the bag-of-visual-words model to mitigate the semantic gap problem in copy-move forgery detection.
Generally, keypoint-based approaches are more efficient and robust against geometric transformations than block-based methods. However, they often perform worse in homogeneous regions, where few or even no keypoint exist.

Recently, deep CMFD frameworks have been attracting increasing attention and showing promising detection results \cite{ferrari_busternet_2018, barni2020copy, zhu_ar-net_2020, zhong_end--end_2020, he2023image, islam_doa-gan_2020, liu2021two, chen_serial_2020, weng2023ucm}. Different from conventional methods, features of deep CMFD frameworks are adaptively learned from a large dataset. 
Wu \textit{et al.} \cite{ferrari_busternet_2018} pioneered the deep CMFD framework with source/target separation. Specifically, they designed a dual-branch framework to localize the similar regions and the forged regions, where identifying similar regions primarily relies on computing the self-correlation of high-level features.
Then, features of these two branches are fused to identify the source and target regions. Different from the work \cite{ferrari_busternet_2018}, Brani \textit{et al.} \cite{barni2020copy} focused only on source/target discrimination, which leverages the irreversibility caused by interpolation artifacts from copy-move transformations and the presence of boundary forgery traces in the target area to disambiguate the source and target regions with the similarity masks obtained by other detection algorithms. Zhu \textit{et al.} \cite{zhu_ar-net_2020} proposed a coarse-to-fine architecture for CMFD. Similar to the work \cite{ferrari_busternet_2018}, they computed the pairwise correlation between feature points in the feature map for locating similar regions. Furthermore, the position attention and the channel attention are adaptively fused to improve detection accuracy. 
 Zhong \textit{et al.} \cite{zhong_end--end_2020} devised a deep CMFD algorithm without source/target separation based on pyramid feature extraction and hierarchical post-processing. Its feature matching method is inspired by the matching approaches used in SIFT \cite{amerini_sift-based_2011}.  Islam \textit{et al.} \cite{islam_doa-gan_2020} suggested identifying copy-move regions using a dual attention mechanism, and they further employ adversarial training to refine the obtained detection masks. Liu \textit{et al.} \cite{liu2021two} designed a proposal superglue to segment objects in the image, thus removing false-alarmed regions and remedying incomplete areas. Motivated by \cite{ferrari_busternet_2018}, Chen \textit{et al.} \cite{chen_serial_2020} proposed to rearrange the parallel architecture proposed in \cite{ferrari_busternet_2018} serially, where the similarity masks were identified using double-level self-correlation, and the result was directly used as the subsequent source/target discrimination. Recently, {Weng \textit{et al.} \cite{weng2023ucm} devised a U-Net-like architecture with multiple asymmetric cross-layer connections leveraging self-correlation for CMFD. It treats large and small tampered regions differently by employing deep backbone networks with semantic information for large regions while lightweight backbone networks for small regions.} Despite the promising results of the above deep CMFD algorithms, they still bear some fundamental limitations as we discussed in Section \ref{sec:intro}. 


\subsection{PatchMatch}\label{sec:intro-PM}
The study of PM was suggested in \cite{barnes2009ToG}, which is an efficient algorithm to find approximate nearest correspondences across different images for structural editing. The original PM has three main procedures. 
In the first step, each pixel of the input image is assigned with a random offset indicating the position of its correspondence in the target image. The rationale of PM is that a large number of random guesses often result in a certain number of optimal or sub-optimal offsets. 
Then, in the second step, those good offsets can be very efficiently propagated to their neighbors since they usually have coherent matches, thus significantly pruning out the search space. In the third step, PM applies a random search to reduce the risk of suboptimality. The last two procedures are often executed several times until convergence. In the past years, several extensions of PM have also been proposed \cite{Hu2016CVPR, Galliani2015ICCV}. Due to its high efficiency, PM has been widely studied in a wide range of applications, such as Stereo Matching \cite{Xu2015TIP},  multi-view stereo \cite{schonberger2016CVPR} and optical flow estimation \cite{Bao2014TIP}. Cozzolino \textit{et al.} \cite{cozzolino_efficient_2015} first employed PM for CMFD, greatly improving the efficiency of conventional block-based algorithms. Note that the standard PM and its extensions are non-differentiable, which cannot be directly integrated into a deep learning framework for end-to-end learning. Recently, Duggal \textit{et al.} \cite{Shivam2019ICCV} developed a differentiable PM module for efficient stereo matching, where they unrolled the standard PM as a recurrent neural network and utilized predefined one-hot filters for offset propagation. 
Motivated by \cite{Shivam2019ICCV}, in this work, we design a new differentiable PM module tailored for CMFD, thus achieving reliable point-to-point matching between source and target regions.

\begin{figure*}[t!]
	\centering
	\includegraphics[width=0.9\linewidth]{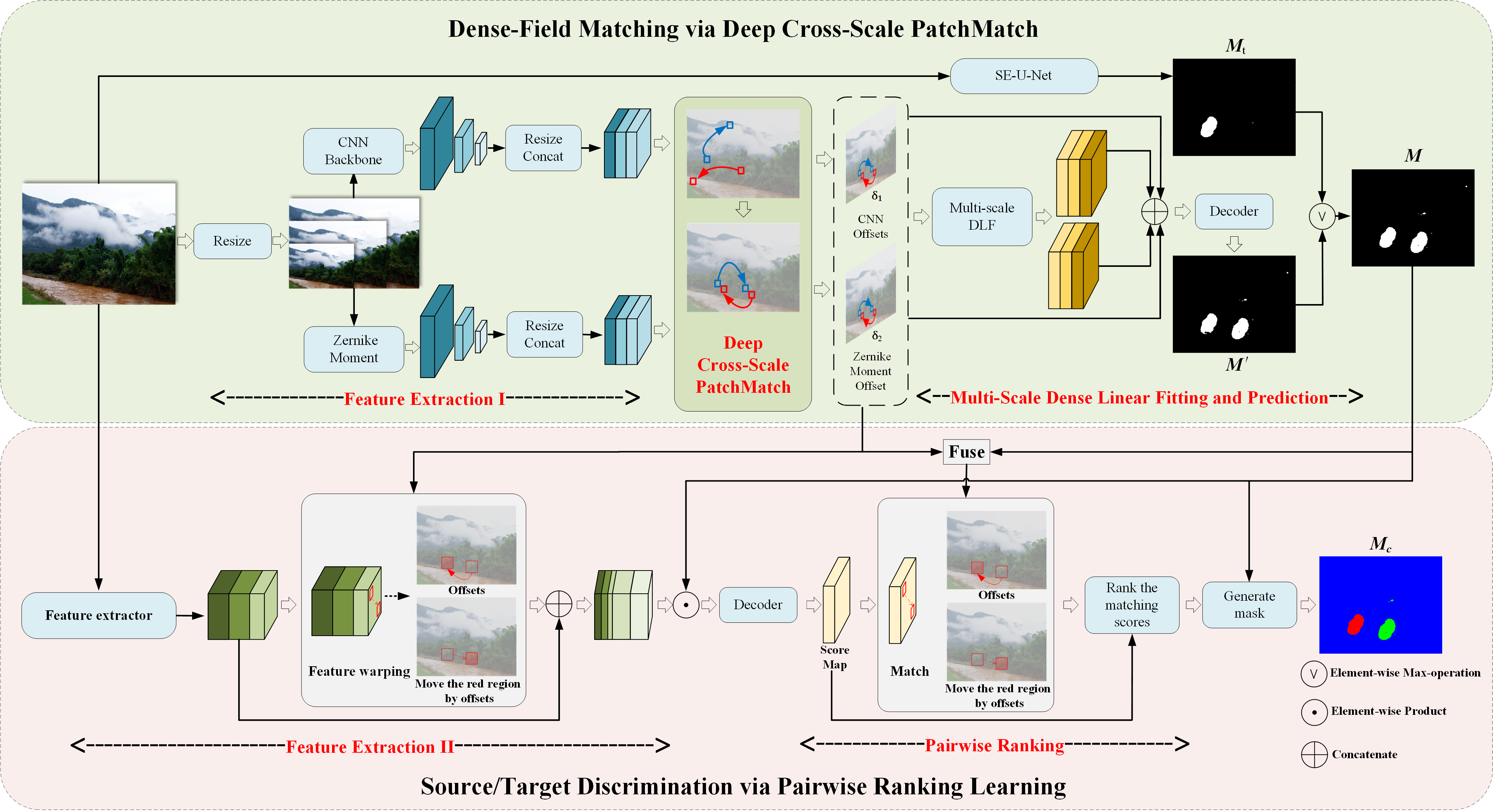}
	\centering
	\caption{The illustration of our framework. The top branch is used to localize copy-move regions via Deep Cross-Scale PM, and the bottom branch is used to differentiate the source and target regions via Pairwise Ranking Learning.}
	\label{fig:framework}
\end{figure*}
\section{Problem Formulation.} \label{sec:problem formulation}
Given a real image, copy-move forgery involves the process of copying a small part of the image and pasting it to a different location within the same image. This technique is commonly used to conceal objects of interest or to duplicate objects in the image.
The objective of CMFD is to identify the areas within an image where the copy-move forgery action has occurred. The area of the image used for copying is referred to as the \textit{source region}, and the area that is covered by the copied region is called the \textit{target region}.

CMFD can be classified into single-channel CMFD and three-channel CMFD based on whether it distinguishes between source and target regions. Single-channel CMFD can be viewed as a pixel-wise binary classification problem, where each pixel is classified as either belonging to the background or the target/source region. On the other hand, three-channel CMFD builds upon single-channel CMFD by further distinguishing between source and target regions, determining whether the detected pixels originate from the source region or the target region. According to convention, we define the following related masks

\begin{itemize}
    \item $\boldsymbol{M}^{gt}$: the ground-truth single-channel mask, where the background region is represented by 0, while pixels for both the source and target areas are set to 1.
    \item$\boldsymbol{M}$: the predicted single-channel mask, where each value reflects the probability that the pixel belongs to the target/source region.
    \item $\boldsymbol{M}_c^{gt}$: the ground-truth three-channel mask, where the background region is represented by [0 0 1] (blue), the source region by [0 1 0] (green), and the target region by [1 0 0] (red).
    \item $\boldsymbol{M}_c$: the predicted three-channel mask.
    \item $\boldsymbol{M}_{t}^{gt}$: the ground-truth mask of the target region, where pixels in the target region are 1, and others are 0.
    \item $\boldsymbol{M}_{t}$: the predicted mask of the target region.
\end{itemize}



\section{The framework of our proposed algorithm.} \label{sec:proposed}
In this section, we discuss our proposed CMFD method via deep PM and pairwise ranking learning. 

\subsection{D2PRL Overview}
As illustrated in Fig.\ref{fig:framework}, the framework of our method has two main branches:
\begin{itemize}
    \item \textbf{Dense-Field Matching via Deep Cross-Scale PM (DFM)}: This branch aims to localize copy-move regions via dense-field matching. By resorting to the designed deep cross-scale PM algorithm, our method seeks explicit and reliable point-to-point matching between source and target regions using features from high-resolution scales.  Compared to previous deep CMFD methods, our detection framework is of high generalizability to various copy-move contents, including objects, incomplete objects and background. 
    \item \textbf{Source/Target Discrimination via Pairwise Ranking Learning (STD)}: This branch utilizes the strong prior knowledge of dense-field matches to differentiate the source and target regions. Based on the matching information, we convert the original discrimination problem into a pairwise ranking problem, which forces the network to reveal slight clues for discriminating the source regions and target region even when the target regions fit well with the background. 
\end{itemize}

\subsection{Dense-Field Matching via Deep Cross-Scale PM (DFM)}
The point-wise relationships serve as a critical clue in image copy-move forgery detection and localization, which have been extensively studied in the conventional CMFD algorithms for reliable decisions. However, the matching procedures of conventional approaches are either non-differentiable or highly complex due to the huge sample space, precluding their usage in modern deep learning frameworks. Therefore, most existing deep CMFD algorithms \cite{ferrari_busternet_2018, islam_doa-gan_2020, chen_serial_2020} calculate an attention map based on pairwise correlation between feature points to present the possibility of a patch is copy-moved, using only high-level features from a very small-sized feature map. As our experiments will demonstrate, this may make them susceptible to overfitting on objects in the training images.

Different from previous works, our DFM branch identifies copy-move regions through point-to-point matching using feature maps of high-resolutions, which effectively combines the merits of both deep and  conventional models. As shown in the top of Fig. \ref{fig:framework}, the DFM branch consists of three blocks, i.e., 1) Feature extraction, 2) Cross-scale matching and 3) error fitting and prediction.

\subsubsection{\textbf{Feature Extraction I}}
The popular feature extraction frameworks, such as VGG and Resnet are designed to progressively reduce the resolutions of feature maps via the pooling operation, thus enlarging the receptive field. Such a strategy is helpful in visual object classification, and has also been utilized in existing deep CMFD models \cite{ferrari_busternet_2018, islam_doa-gan_2020, chen_serial_2020}. However, it may not be effective for CMFD, since the resulting networks could be easily overfitted to objects in the training images. In contrast to previous deep models, the DFM branch aims to extract high-resolution features, making our framework highly generalizable to non-objects.

The feature extraction of our DFM branch consists of a pyramid feature learning block and a ZM extraction block. In practice, the target region can be heavily resized before pasting. Motivated by the conventional feature extraction algorithm SIFT \cite{Lowe04}, we propose a pyramid feature extraction to obtain multi-level features to fight against severe rescaling attacks. Specifically, we first downsample  or upsample the given image ${I}_{o}\in{R}^{H\times{W}\times{3}}$ by
\begin{equation}
    I_b = resize(I_o, r_b), ~~I_u=resize(I_o,r_u),
\end{equation} where $r_b$ and $r_u$ are set to $0.75$ and $1.5$ in our experiment. The obtained images $I_b, I_o$ and $I_u$ of different resolutions are then fed into a backbone network for feature learning.  
Our backbone architecture comprises five convolution blocks, each consisting of a convolution layer, a BatchNorm layer, and a ReLU layer. At the end of the backbone, there is a resizing layer to rescale the feature maps to the same dimensions of $H\times W\times c$, where $c$ represents the number of feature channels, empirically set to 32. Finally, we obtain three feature maps as  
\begin{equation}\label{eq:mcalefeature}
    \boldsymbol{F}_u = \mathcal{F}_{et}(I_u), \boldsymbol{F}_o = \mathcal{F}_{et}(I_o), ~\boldsymbol{F}_b = \mathcal{F}(I_b),
\end{equation} where $\mathcal{F}_{et}(\cdot)$ is the feature extraction function. Note that we do not use pooling layers in the DFM branch for maintaining high resolution feature maps.

During our experiment, we observed that training a pure CNN feature extraction from random initialization proves challenging in acquiring rotation-invariant features for the CMFD task. To address this, we also incorporate conventional ZM \cite{teague1980image} as a complement to CNN features, enhancing robustness against rotations. Given an image $I(x,y)$, where $(x,y)\ \in\ R^2$, its representation in the polar coordinates space can be written as $I(\rho,\theta)$, with $\rho\ \in\ [0, 1]$ and $\theta \in [0, 2\pi]$. The ZM of order $p$ with repetition $q$ for ${I}(\rho,\theta)$ is defined as 
\begin{equation} \label{eq:ZMC}
		\begin{aligned}
		\boldsymbol{F}_{ZM}(p,q) =\frac{n+1}{\pi} \int_{0}^{2\pi}\int_{0}^{1}I(\rho,\theta)K^*_{p,q}(\rho,\theta){\rho}d{\rho}d{\theta}.
		\end{aligned}
	\end{equation} 
 Here, $K^*_{p,q}(\rho,\theta)$ denotes the complex conjugate of Zernike polynomial $K_{p,q}(\rho,\theta)$, which is given by 
\begin{equation}
    K_{p,q}(\rho,\theta) = R_{p,q}(\rho)exp(jm\theta),
\end{equation} where the orthogonal radial polynomial $R_{p,q}(\rho)$ is defined as
\begin{equation}
    R_{p,q}(\rho) = \sum_{s=0}^{(1-|p|)/2}\frac{(-1)^s[(1-s)!]\rho^{1-2s}}{s!
    \left(\frac{1+|p|}{2}-s\right)!\left(\frac{1-|p|}{2}-s\right)!}.
\end{equation} 
Note that image rotation does not change the magnitude of $\boldsymbol{F}_{ZM}(p,q)$, and thus ZM are rotation invariance \cite{Xin2007TIP}. 
For digital images, we use summations to approximate integrals, and (\ref{eq:ZMC}) can be written as \begin{equation}
		\begin{aligned} \label{eq:ZM}
		\boldsymbol{F}_{ZM}(p,q) = \frac{p+1}{\pi}\sum_{(x,y)\in{\Omega_{xy}}}I(x,y)K^*_{p,q}(\rho_{xy},\theta_{xy}), 
		\end{aligned}
\end{equation} where $\rho_{xy} =\sqrt{x^{2}+y^{2}}$, $\theta_{xy} = {arctan^{-1}(y/x)}$, and $\Omega_{xy}$ is a set of pixels centered by $(x,y)$. Note that the feature extraction process (\ref{eq:ZM}) can be implemented using a convolution layer, where the filters are computed by $K^*_{p,q}(\rho_{xy},\theta_{xy})$. 

In our work, we set the maximum order of ZM to 5, which yields a 12-dimensional feature map for the input image. Then, we can obtain three feature maps of $H\times{W}\times{12}$ for ${I}_u, {I}_o$ and ${I}_d$ as
\begin{equation}
    \boldsymbol{F}'_u = \mathcal{F}_{zm}(I_u), ~\boldsymbol{F}'_o = \mathcal{F}_{zm}(I_o), ~\boldsymbol{F}'_b = \mathcal{F}_{zm}(I_b).
\end{equation}

\subsubsection{\textbf{Deep Cross-Scale PM}} The identification of reliable pixel matches holds significant importance in copy-move forgery detection. For an image $I \in \mathbb{R}^{H\times W\times 3}$, we define $\boldsymbol{\delta} \in \mathbb{R}^{H\times W\times 2}$ as an offsets map determined by the nearest-neighbor field over all possible pixel coordinates. Mathematically, 
\begin{equation}\label{eq:offset}
	\begin{aligned}
	    \boldsymbol{\delta}(i,j) =\arg\min_{i_s,j_s\neq 0} Dis(\boldsymbol{F}(i,j),\boldsymbol{F}(i+i_s,j+j_s)),
	\end{aligned}
\end{equation} where $0 \leq i+i_s\leq H$, $0 \leq j+j_s\leq W$, $\boldsymbol{F}(i,j)$ denotes the feature vector at location $(i,j)$, and $Dis(\cdot)$ represents an appropriate distance metric.  Clearly, finding dense pixel matches corresponds to solving the optimization problem (\ref{eq:offset}). For simplicity, we denote $\boldsymbol{\delta}(i,j)$, i.e., the solution to the problem (\ref{eq:offset}), as $(\boldsymbol{\delta}_{ij}^x, \boldsymbol{\delta}_{ij}^y)$, and 
define 
\begin{equation}
    match(i,j) = (i+\boldsymbol{\delta}_{ij}^x,j+\boldsymbol{\delta}_{ij}^y),
\end{equation} as a function that finds the nearest neighbor in the feature space for the point $(i, j)$. 
The PM algorithm efficiently provides an approximate solution to the problem (\ref{eq:offset}) by leveraging the spatial regularity present in natural images. The classic PM consists of three main procedures as detailed in Section \ref{sec:intro-PM}, i.e.,  i) offset initialization, ii) offset propagation and 3) random search. Despite its efficiency, the classic PM method heavily relies on manually designed features and cannot be directly integrated into deep learning frameworks due to its non-differentiability. 

\begin{figure*}[t!]
	\centering
	\includegraphics[width=0.9\linewidth]{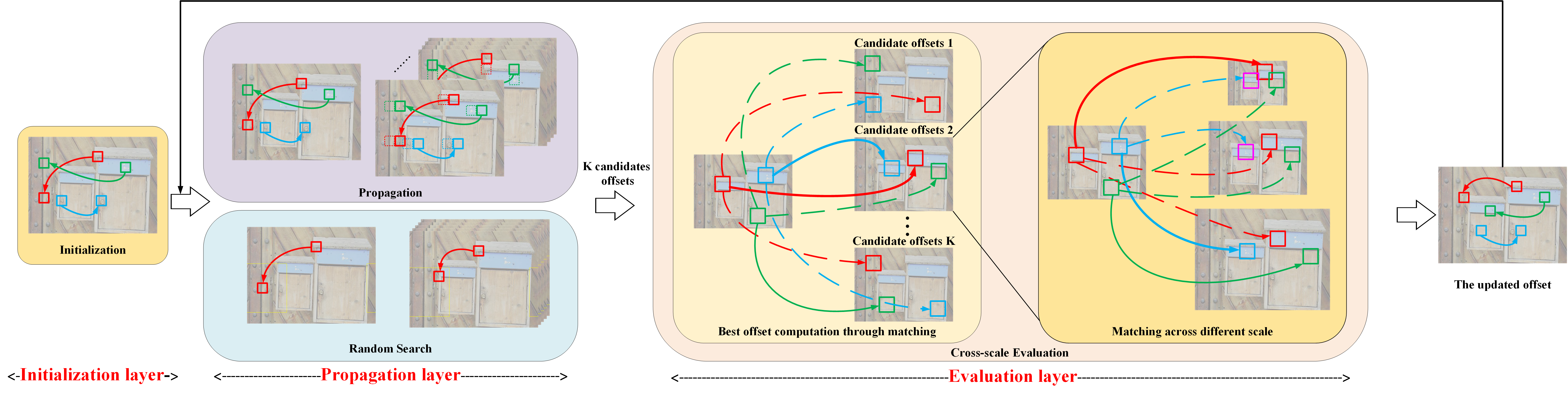}
	\centering
	\caption{The framework of our deep cross-scale PM. The Initialization layer first generates a valid offset for each pixel. Then, the Propagation layer uses propagation and random search to generate $K$ candidate offsets for each pixel. In the Evaluation layer, for each offset, the optimal matching score is calculated across different scale feature maps; the best offset is chosen from the $K$ candidate offsets for each pixel (indicated by solid arrows), and the corresponding matching score is saved. Finally, the offset with the highest matching score is then passed back to the Propagation layer.}
	\label{fig:patchmatch}
\end{figure*}

In this study, we devise a differentiable deep cross-scale PM module tailored for CMFD, aiming to establish point-to-point matching between source and target regions. The framework of our cross-scale PM is illustrated in Fig. \ref{fig:patchmatch}. Our design is inspired by the work \cite{Shivam2019ICCV}, which suggests a differentiable PM module for stereo matching. However, due to distinct tasks, our framework differentiates from \cite{Shivam2019ICCV} in many aspects. First, the method presented in \cite{Shivam2019ICCV} is limited to narrow-range horizontal matching between two nearly identical images captured from slightly different viewpoints. In contrast, we design our PM method to operate within a single image where only parts of the regions (potentially over long ranges) are duplicated. Secondly, the work \cite{Shivam2019ICCV} focused solely on narrow-range horizontal matching, without explicitly addressing rotation or scaling problems that may occur in copy-move forgery. Additionally, we propose a cross-scale evaluation layer to address large rescaling transformations. Note that offsets obtained by \cite{Shivam2019ICCV} are used for the disparity range estimation, while we use the offsets to compute a forgery heat map through multi-scale dense linear fitting. 
The details descriptions of our cross-scale PM are given below.


i) Initialization layer: Initially, we generate random offsets $\boldsymbol{\delta}(i,j)$ for each pixel $I(i,j)$, ensuring that $(i+\boldsymbol{\delta}_{ij}^x,j+\boldsymbol{\delta}_{ij}^y)$ remains a valid location within the image. We exclude $\boldsymbol{\delta}(i,j)=(0,0)$ to eliminate trivial solutions. Note that with a substantial number of random offsets, it is highly probable that some will be optimal or close to optimal.

ii) Propagation layer: The purpose of this layer is to propagate pixel offsets to their neighboring pixels. Due to the fact that natural images are locally coherent, adjacent pixels usually have similar offsets, and then those good offsets can be quickly propagated to their neighbors. The propagation layer comprises a propagation block and a random search block.

\begin{figure}	
	\centering
	\begin{minipage}[t]{0.5\linewidth}
        \centering
        \includegraphics[width=\textwidth]{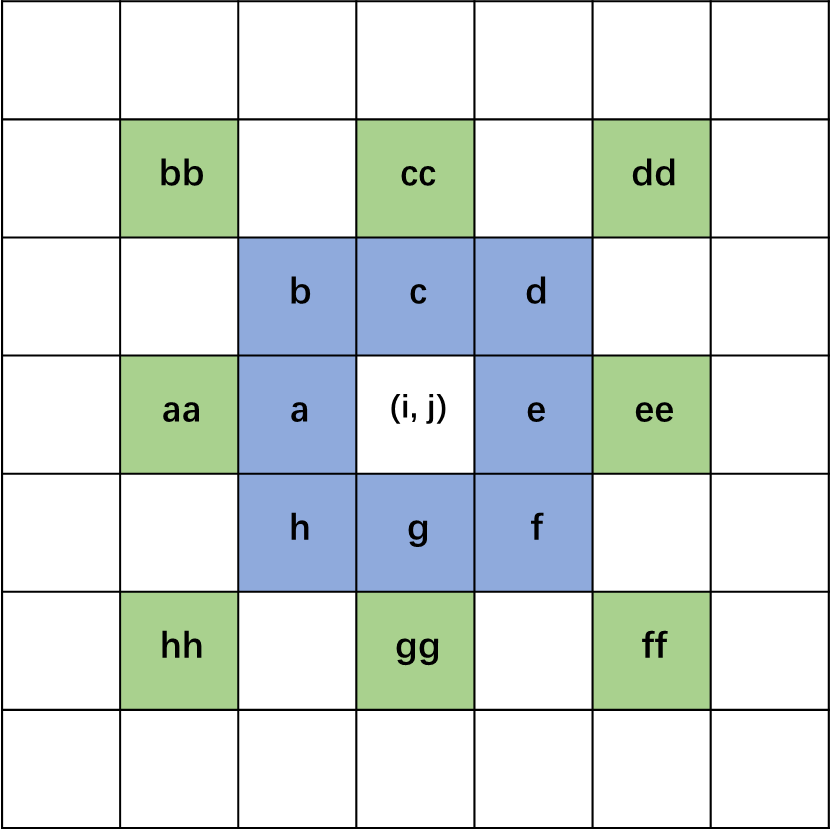}
    \end{minipage}
\caption{Pixels used for propagation. Different letters mark the relative coordinates centered around $(i, j)$. Green and blue pixels propagate their offsets to pixel $(i, j)$. Blue pixels propagate directly, while green pixels use first-order predictors for propagation.}\label{fig:predictors}
\end{figure}
\begin{figure}	
	\centering
	\begin{minipage}[t]{0.75\linewidth}
        \centering
        \includegraphics[width=\textwidth]{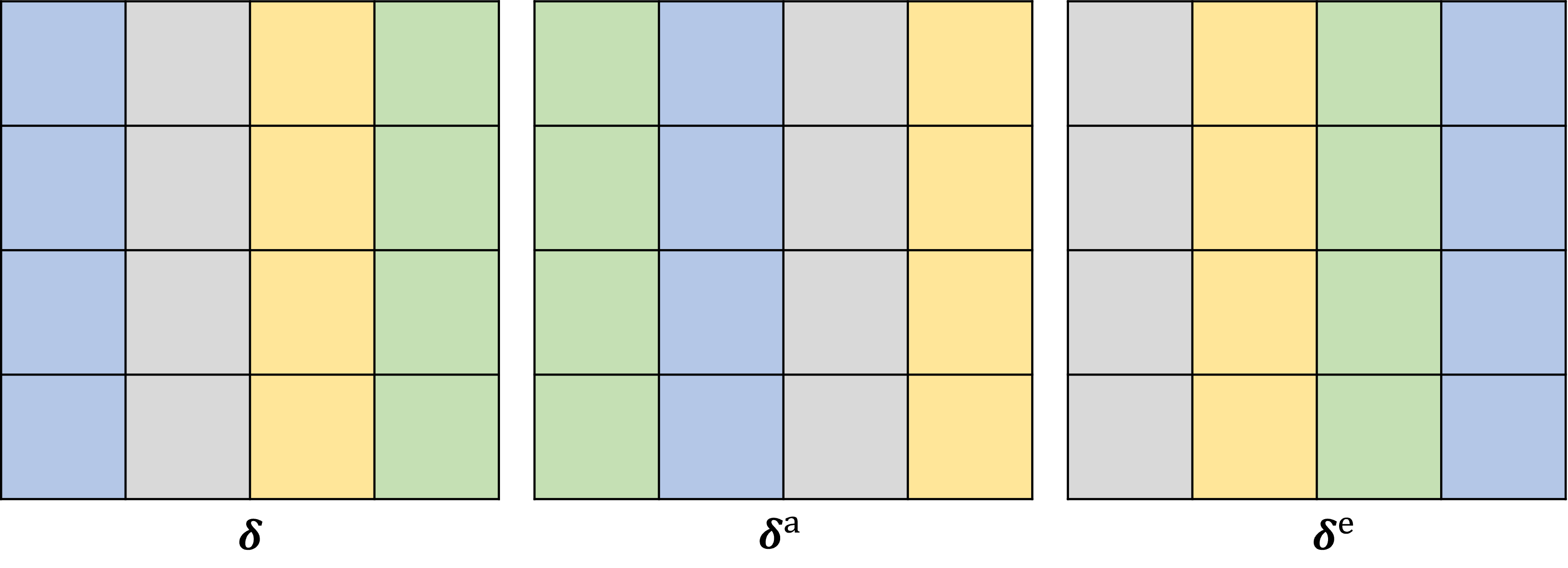}
    \end{minipage}
\caption{
Visualization examples demonstrating the propagation of circular shift displacement to generate candidate offset maps $\boldsymbol{\delta}^a$ and $\boldsymbol{\delta}^e$ based on $\boldsymbol{\delta}$, where each different color represents that the offsets belong to the same column rather than having the same offset.}\label{fig:circular_shift}
\end{figure}

In our work, we refer to the propagation that utilizes only offsets from the most adjacent pixels as the zero-order propagation. As Illustrated in Fig. \ref{fig:predictors}, four zero-order candidate offsets for each pixel are created as follows 
\begin{equation}\label{eq:zero-order}
		\begin{aligned}
		\boldsymbol{\delta}^{\gamma}(i,j) &= \boldsymbol{\delta}(i^{\gamma},j^{\gamma}),\\
		\gamma &\in \{a, c, e, g\},
		\end{aligned}
\end{equation}
where ($i^{\gamma},j^{\gamma}$) denotes pixel coordinates shown in Fig. \ref{fig:predictors}. We implement the propagation procedure defined in Eq. (\ref{eq:zero-order}) using the circular shift operation. Fig. \ref{fig:circular_shift} illustrates the propagation to generate candidate offset maps $\boldsymbol{\delta}^a$ and $\boldsymbol{\delta}^e$.
Note that zero-order propagation is effective in addressing CMFD with rigid translations; however, its efficacy diminishes when dealing with rotation and rescaling transformations, which are common operations used in copy-move forgeries. Inspired by \cite{cozzolino_efficient_2015}, in this study, we also consider the first-order propagation, which leverages the offset information of two consecutive pixels in the same direction. To be specific, we additionally generate eight first-order candidate offsets for each pixel using the following approach
\begin{equation}
		\begin{aligned}
		\boldsymbol{\delta}^{\gamma\gamma}(i,j) &= 2\boldsymbol{\delta}(i^{\gamma},j^{\gamma})-\boldsymbol{\delta}(i^{\gamma\gamma},j^{\gamma\gamma}),\\
		\gamma &\in \{a, b, c, d, e, f, g, h\}.
		\end{aligned}
\end{equation}

To prevent being trapped in local optima, we further apply a random search for each pixel. specifically, four more candidate offsets for each pixel, denoted by $\boldsymbol{\delta}^{r1}(i,j), \boldsymbol{\delta}^{r2}(i,j), \boldsymbol{\delta}^{r3}(i,j)$ and $\boldsymbol{\delta}^{r4}(i,j)$, are randomly generated as below
\begin{equation}
    \begin{aligned}
    \boldsymbol{\delta}^{\gamma}(i,j) &= \boldsymbol{\delta}(i,j) + \Delta \boldsymbol{\delta}(\gamma),\\
    \gamma &\in \{r1, r2, r3, r4\}.
    \end{aligned}
\end{equation} where $\Delta \boldsymbol{\delta}(\cdot)$ is randomly sampled from a predefined search space centered at $(i,j)$. We empirically set the radius of the search region to 50 in our study.

By applying the aforementioned propagation processes, we ultimately generate seventeen potential offsets for each pixel
\begin{equation}
    \boldsymbol{\delta}_c = \{{\boldsymbol{\delta}}, \boldsymbol{\delta}^{a},..., \boldsymbol{\delta}^{g}, \boldsymbol{\delta}^{aa},...,\boldsymbol{\delta}^{hh},\boldsymbol{\delta}^{r1},..., \boldsymbol{\delta}^{r4}\},
\end{equation} where $\boldsymbol{\delta}_c$ represents the set of candidate offset maps, including four candidate offset maps through zero-order propagation, eight candidate offset maps through first-order propagation, four candidate offset maps through random search, and one offset map $\boldsymbol{\delta}$ output from the previous round of the evaluation layer or the initial offset map in the first round of propagation. To simplify, we write $\boldsymbol{\delta}_c = \{\boldsymbol{\delta}^1, \boldsymbol{\delta}^2,...,\boldsymbol{\delta}^K\}$, where $K=17$.

iii) Evaluation layer: The evaluation layer is devised to determine the optimal offset for each pixel from the $K$ candidate offsets obtained in the propagation layer.  For simplicity, we employ  $\ell_1$-norm to quantify the difference in the distance between two feature vectors. Thus, the matching score for a given offset $\boldsymbol{\delta}^{k}(i,j)$ is calculated as follows:
\begin{equation}\label{eq:SK}
	\begin{aligned}
		\boldsymbol{S}^{k}(i,j) = -\left|\left|f(i,j)- f\left((i,j)+ \boldsymbol{\delta}^{k}(i,j))\right)\right|\right|_1.
	\end{aligned}
\end{equation}

where $f(\cdot)$ denotes a function to extract the feature vector of a given location in the feature map.
Ideally, we can apply $argmax$ function to obtain the best offset with the largest matching score $\boldsymbol{S}^{k}(i,j)$. However, the $argmax$ function is non-differentiable, which precludes gradient back-propagation during network training. According to \cite{Kendall_2017_ICCV}, we adopt a relaxed version of the $argmax$ function to compute the best offset
\begin{equation}\label{eq:hat_S}
	\begin{aligned}
		\boldsymbol{\delta}^{p}(i,j) = \sum_{k=1}^K\boldsymbol{\delta}^{k}(i,j)\times{\boldsymbol{\hat{S}}^{k}(i,j)},
	\end{aligned}
\end{equation}where 
\begin{equation}
    \boldsymbol{\hat{S}}^{k}(i,j) = \frac{\exp(\beta \boldsymbol{S}^{k}(i,j))}{\sum_{k=1}^{K}\exp(\beta  \boldsymbol{S}^{k}(i,j))},
\end{equation} where $\beta$ is a temperature parameter.  
Subsequently, $\boldsymbol{\delta}^{p}$ is reintroduced into the propagation layer, serving as the initial offset map $\boldsymbol{\delta}$ for the next iteration. It should be emphasized that $\boldsymbol{\delta}^{k}(i,j)$ is from a continuous space (see Eq. \ref{eq:hat_S}). Therefore, we cannot directly obtain the corresponding feature using coordinates. In this work, we employ bilinear interpolation, which is differentiable, to obtain the corresponding feature vector $f((i,j)+ \boldsymbol{\delta}^{k}(i,j))$.



\begin{figure}
	\centering
	\includegraphics[width=1\linewidth]{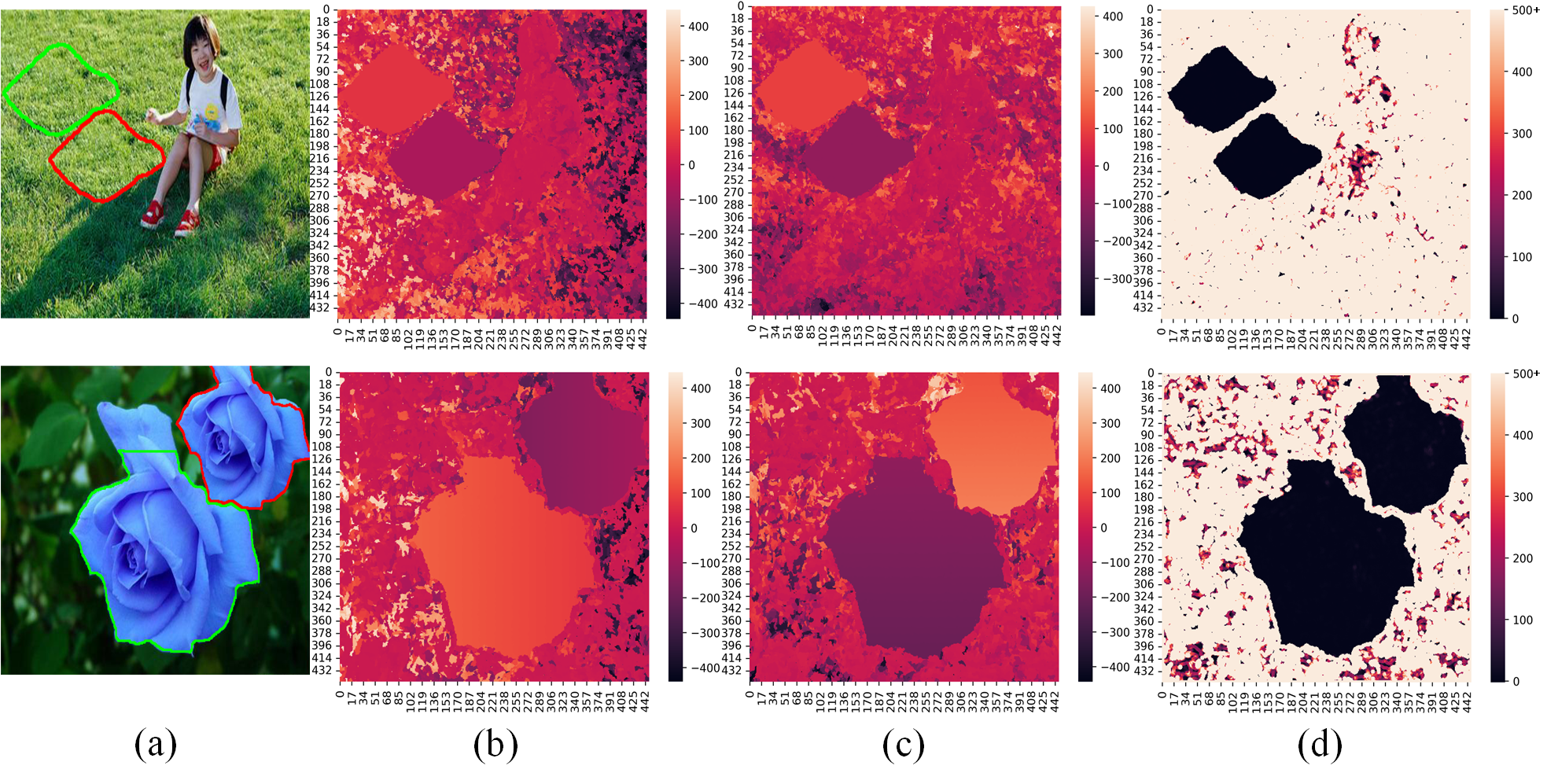}
	\centering
    \caption{Visualization of offset maps and DLF maps. (a) copy-move images; (b) offset maps $\boldsymbol{\delta}_1$ ($x$ coordinate); (c) offset maps $\boldsymbol{\delta}_1$ ($y$ coordinate); (d) DLF maps with diameter $\rho = 7$.}
    \label{fig:dlf}
\end{figure}

In order to handle large rescaling transformations in copy-move forgeries, we further propose a cross-scale PM algorithm as shown in Fig. \ref{fig:patchmatch}. In contrast to the conventional approach, we evaluate offsets across features of different scales. Concretely, considering the CNN features $\boldsymbol{F}_u, \boldsymbol{F}_o$ and $\boldsymbol{F}_b$  obtained by Eq. (\ref{eq:mcalefeature}), we calculate the matching scores as follows:
\begin{equation}
	\begin{aligned}
		\boldsymbol{S}_{nm}^{k}(i,j) = -&\left|\left|f_n(i,j)- f_m\left((i,j)+ \boldsymbol{\delta}^{k}(i,j)\right)\right|\right|_1.\\
		&~~~~~~~~~~~~~~~~~~n,m\in\{u, o, b\},
	\end{aligned}
\end{equation}
where $f_n(\cdot)$ and $f_m(\cdot)$ extract the feature vector of a given location based on the feature maps $\boldsymbol{F}_n$ and $\boldsymbol{F}_m$, respectively.
We retain only the best cross-scale matching score for each candidate offset, and $\boldsymbol{S}^{k}(i,j)$ in Eq. (\ref{eq:SK}) can be computed as 
\begin{equation}
	\boldsymbol{S}^{k}(i,j) = \max(\boldsymbol{S}_{nm}^{k}(i,j)), ~n,m\in\{u, o, b\}.
\end{equation} Note that the propagation layer and the evaluation layer ensure that previously found good offsets quickly propagate to neighboring pixels. These two layers will be recurrently executed several times to achieve good offsets for all the pixels.  
To simplify, we define the resultant offset map from CNN features as $\boldsymbol{\delta}_1$, and the offset map from ZM features as $\boldsymbol{\delta}_2$. Figs. \ref{fig:dlf}(b-c) illustrate two examples of the resulting offset maps.

\subsubsection{\textbf{Multi-scale dense linear fitting and prediction}}
In this subsection, we discuss how to reveal the potential copy-move regions using the obtained offset maps $\delta_1$ and $\delta_2$.
Ideally, the offset map should exhibit smooth linear behavior in copy-move regions, while chaotic in other regions as shown in Figs. \ref{fig:dlf}. Recall that the offset establishes the matching relationship between two pixels. One can use Random Sample Consensus (RANSAC)\cite{fischler1981random} technology to filter out those correct matches. However, the conventional RANSAC algorithm is non-differentiable and very time-consuming. 
Inspired by \cite{cozzolino_efficient_2015}, we propose to evaluate the matching quality based on dense linear fitting (DLF).

Let
\begin{equation}
   \boldsymbol{P} = \begin{bmatrix}
     p_{11}, &p_{12}, &1\\
    p_{21}, &p_{22}, &1 \\
    ~&\vdots &~\\
    p_{N1}, &p_{N2}, &1 
    \end{bmatrix}
\end{equation} be homogeneous coordinates of $N$ pixels centered at $(p_{N1/2}, p_{N2/2})$. If these $N$ pixels are correctly matched with their cloned correspondences, then there exists an affine transformation such that
\begin{equation}
   \boldsymbol{P} + \boldsymbol{\delta}_P \approx \boldsymbol{P}\boldsymbol{A}.
\end{equation} Here, $\boldsymbol{A}$ represents the affine transformation matrix, and $\boldsymbol{\delta}_P$ contains the offsets of $\boldsymbol{P}$ obtained through PM algorithm.  Let $\boldsymbol{B} = \boldsymbol{A}-\boldsymbol{I}$, then the optimal $\boldsymbol{B}$ can be achieved by solving the following linear regression problem
\begin{equation}\label{eq:B}
    \boldsymbol{B}^* = \arg\min_{\boldsymbol{B}}||\boldsymbol{{\delta}}_P - \boldsymbol{P}\boldsymbol{B}||_F^2.
\end{equation} 
The above problem has a closed-form solution
\begin{equation}
    \boldsymbol{B}^* = (\boldsymbol{P}^T\boldsymbol{P})^{-1}\boldsymbol{P}^T\boldsymbol{\delta}_P.
\end{equation} Note that optimizing the $x$-component and the $y$-component of $\boldsymbol{B}$ is independent. The fitting error for the $x$-component can be computed as 
\begin{equation}\label{eq:errx}
\begin{aligned}
     \epsilon_x^2 &= ||\boldsymbol{\delta}_x -  \boldsymbol{P}(\boldsymbol{P}^T\boldsymbol{P})^{-1}\boldsymbol{P}^T\boldsymbol{\delta}_x||^2,\\
     &=\boldsymbol{\delta}_x^T(\boldsymbol{I}-\boldsymbol{H})\boldsymbol{\delta}_x,
\end{aligned}
\end{equation} where $\boldsymbol{H} =  \boldsymbol{P}(\boldsymbol{P}^T\boldsymbol{P})^{-1}\boldsymbol{P}^T$, and $\boldsymbol{\delta}_x\in R^{N}$ denotes the $x$-component of $\boldsymbol{\delta}_P$. The second equation of (\ref{eq:errx}) holds since $\boldsymbol{H}$ is symmetric and idempotent. We can further decompose $\boldsymbol{H}$ as 
\begin{equation}
    \boldsymbol{H} = \boldsymbol{Q}\boldsymbol{Q}^T, ~~~\boldsymbol{Q} = [\boldsymbol{q}_1, \boldsymbol{q}_2, \boldsymbol{q}_3]
\end{equation} where $\boldsymbol{Q}$ an orthogonal matrix, and $\boldsymbol{q}_i \in \boldsymbol{R}^N$. Then, we can rewrite (\ref{eq:errx}) as 
\begin{equation}
    \epsilon_x^2 = (\boldsymbol{\delta}_x^T\boldsymbol{\delta}_x)-(\boldsymbol{\delta}_x^T\boldsymbol{q}_1)^2-(\boldsymbol{\delta}_x^T\boldsymbol{q}_2)^2-(\boldsymbol{\delta}_x^T\boldsymbol{q}_3)^2.
\end{equation} Similarly, we can compute the fitting error for the $y$-component, which is denoted as $\epsilon_y^2$.
Finally, we obtain the fitting error of the pixel $(p_{N1/2}, p_{N2/2})$  as
\begin{equation}
    \epsilon^2 = \epsilon_x^2 + \epsilon_y^2.
\end{equation}

Note that choosing a suitable $N$ is important for an accurate estimation of $\epsilon^2$. A large $N$ may include many non-copy-move pixels, while a small $N$ is noise-sensitive.  Both cases would reduce the estimation accuracy. In practice, we generally have no prior knowledge of the shape and the size of copy-move regions. To address this issue, we propose to estimate $\epsilon^2$ in a multi-scale manner. 
In this work, we consider three diameters $\rho \in  \{7, 9, 11\}$, which yields three DLF maps $\boldsymbol{\epsilon}_1^2,\boldsymbol{\epsilon}_2^2$ and $\boldsymbol{\epsilon}_3^2$.

Subsequently, the DLF maps and offsets are input to a simple decoder to predict a single-channel copy-move mask $\boldsymbol{M}' \in R^{H \times W \times 1}$
\begin{equation}\label{eq:pred}
    \boldsymbol{M}' = Decoder\left(\boldsymbol{\epsilon}_1^2,\boldsymbol{\epsilon}_2^2,\boldsymbol{\epsilon}_3^2, \boldsymbol{\delta}_1, \boldsymbol{\delta}_2\right).
\end{equation} Our decoder comprises five simple convolutional blocks. The first four blocks include a convolution layer with BatchNorm and ReLU activation, while the final block is followed by a sigmoid function. 

Note that the above matching strategy relies solely on spatial similarities between the source and target region features to find copy-paste regions, ignoring inconsistent information between the target region and the background. As a result, it lacks precision in localizing the target regions. In this work, we further refine the target regions of the mask $\boldsymbol{M}$ by exploring the inconsistent information between the target region and background. We achieve this by employing the ``SE-U-net" proposed in \cite{wu2022robust} to generate the potential target region mask $\boldsymbol{M}_t$. The final refined single-channel copy-move mask $\boldsymbol{M}$ is then derived through an element-wise max operation between $\boldsymbol{M}_t$ and $\boldsymbol{M}'$.

\subsection{Source/Target Discrimination via Pairwise Ranking Learning (STD)}
Existing deep CMFD algorithms primarily utilize convolution operations across the entire input image to distinguish between source and target regions, which perform poorly when target regions blend well with the background. Unlike previous deep CMFD approaches, our framework leverages the explicit point-to-point matching from our PM algorithm, establishing a strong guideline for source/target separation: \textit{if a copy-move point $(i, j)$ belongs to the source (target) region, its matching correspondence $(i + \boldsymbol{\delta}_{ij}^x, j + \boldsymbol{\delta}_{ij}^y)$ must be in the target (source) region}. This compelling prior empowers us to effectively differentiate between source and target regions using these matching relationships, even when they are overlapped.

In this work, we consider an equivalent comparison problem, i.e., determining \textit{which point between $(i,j)$ and $(i+\boldsymbol{\delta}_{ij}^x,j+\boldsymbol{\delta}_{ij}^y)$ is more likely to be a cloned pixel}. To ensure end-to-end training of our framework, we propose a pairwise rank learning method for solving the problem as illustrated in Fig. \ref{fig:framework}. 
\subsubsection{\textbf{Feature Extraction II}} First, we employ a simple backbone network for extracting deep features from the input image. Our backbone consists of three blocks, each comprising two convolution layers followed by a maxpool layer. After resizing and concatenating the features from each block, we obtain a feature map denoted as $\boldsymbol{F}_d \in R^{h \times w \times c_1}$, where $h = H/8$, $w = W/8$, and $c_1 = 448$.
Then, we warp the feature $\boldsymbol{F}_d$ as $\boldsymbol{F}_d'$ and  $\boldsymbol{F}_d''$  according to the matching results ($\boldsymbol{\delta}_1$ and $\boldsymbol{\delta}_2$) of our PM algorithm, which can be written as 
\begin{equation}
    \boldsymbol{F}_d'(i,j) = f_d(i+\boldsymbol{\delta}_{1ij}^x,j+\boldsymbol{\delta}_{1ij}^y), 
\end{equation}
\begin{equation}
    \boldsymbol{F}_d''(i,j) = f_d(i+\boldsymbol{\delta}_{2ij}^x,j+\boldsymbol{\delta}_{2ij}^y), 
\end{equation}
where $f_d$ computes the feature vector of a given location using bilinear interpolation based on $\boldsymbol{F}_d$. We concatenate $\boldsymbol{F}_d$, $\boldsymbol{F}_d'$ and $\boldsymbol{F}_d''$  as a new feature map 
\begin{equation}
    \boldsymbol{F}_{st}' = Cat(\boldsymbol{F}_d, \boldsymbol{F}_d', \boldsymbol{F}_d''). 
\end{equation} Apparently, after the warping process, each copy-move pixel of $\boldsymbol{F}_{st}'$ has both the source feature and target feature. 
\subsubsection{\textbf{Pairwise Ranking}}We further use the mask $\boldsymbol{M}$ to reduce the effect of those background pixels
\begin{equation}
    \boldsymbol{F}_{st} = \boldsymbol{F}_{st}' \odot \boldsymbol{M},
\end{equation} where $\odot$ is the element-wise product operation. 
The feature $\boldsymbol{F}_{st}$ is sent to a decoder consisting of five convolution layers and a sigmoid activation function to compute a score map $\boldsymbol{S}_{f} \in{R}^{h\times{w}}$. Let $\boldsymbol{M}_b$ represent the binary result of $\boldsymbol{M}$. 
For each pixel $(i, j)$ satisfying $\boldsymbol{M}_b(i, j) = 1$, $\boldsymbol{S}_{f}(i, j)$ being closer to 1 indicates a higher probability of the pixel belonging to the source region.

After having the score map, we propose to discriminate the matched points through pairwise ranking. Specifically, we compute
\begin{equation}\label{eq:sRank}
		\begin{aligned}
        \boldsymbol{S}_{rank}(i,j) =&\boldsymbol{S}_{f}(i,j) - \boldsymbol{S}_{f}(i+\boldsymbol{\delta}_{fij}^x,j+\boldsymbol{\delta}_{fij}^y).
		\end{aligned}
	\end{equation}
For each pixel $(i, j)$ satisfying $\boldsymbol{M}_b(i, j) = 1$, $\boldsymbol{S}_{rank}(i, j) > 0$ indicates that pixel $(i, j)$ is more likely to belong to the source region, whereas $\boldsymbol{S}_{rank}(i, j) < 0$ suggests it is more likely to belong to the target region. 
It is worth noting that in source/target discrimination, we are only interested in pixels that belong to the source or target region, and 
do not impose explicit constraints on the background pixels, i.e, pixels satisfying $\boldsymbol{M}_b(i, j) = 0$. As shown in Fig. \ref{fig:dlf},  the offsets of background pixels are unordered, and their corresponding elements in $\boldsymbol{S}_{rank}$ can take any value. 
$\boldsymbol{\delta}_{fij}$ in Eq. (\ref{eq:sRank}) is the fused offset of the pixel $(i,j)$ computed from  $\boldsymbol{\delta}_{1}$ and $\boldsymbol{\delta}_{2}$, where $\boldsymbol{\delta}_{1}$ is obtained using CNN features while $\boldsymbol{\delta}_{2}$ is obtained using ZM features. As previously discussed, our framework supplements ZM features with CNN features to acquire rotation-invariant capabilities. Therefore, in our work, when the confidence of the offset at point $(i, j)$ in $\boldsymbol{\delta}_{1}$ is high, the fused offset $\boldsymbol{\delta}_{fij}$ directly adopts the offset value from $\boldsymbol{\delta}_{1}$, i.e., $\boldsymbol{\delta}_{fij}=\boldsymbol{\delta}_{1ij}$; otherwise, $\boldsymbol{\delta}_{fij}$ adopts the result from $\boldsymbol{\delta}_{2}$ to compensate for the limitations of CNN features.
Specifically, we set
\begin{equation}
		\boldsymbol{\delta}_{fij}=
        \begin{cases}
        \boldsymbol{\delta}_{1ij} &\boldsymbol{M}_b(i,j)=1\ and\ \boldsymbol{M}_b(i+\boldsymbol{\delta}_{1ij}^x,j+\boldsymbol{\delta}_{1ij}^y)=1 \\
        \boldsymbol{\delta}_{2ij} &otherwise
        \end{cases}.
\end{equation}
Note that $\boldsymbol{M}_b(i,j)=1$ and $\boldsymbol{M}_b(i+\boldsymbol{\delta}_{1ij}^x,j+\boldsymbol{\delta}_{1ij}^y)=1$ indicate that the two matched points $(i,j)$ and $(i+\boldsymbol{\delta}_{1ij}^x,j+\boldsymbol{\delta}_{1ij}^y)$ are either from the source region or the target region. Therefore, we can consider that these two points are matched correctly with high confidence and set $\boldsymbol{\delta}_{fij}=\boldsymbol{\delta}_{1ij}$. However, for matches that $\boldsymbol{\delta}_{1}$ fails to accurately detect (e.g., pixels after rotation), we rely on the results from $\boldsymbol{\delta}_{2}$ instead.
It should be emphasized that we are not interested in the offset values in background regions, as they will be filtered out when computing the final three-channel mask. 
During the training phase, $\boldsymbol{S}_{rank}$ is supervised through a discrimination loss, which will be elaborated in Section \ref{sec:loss}.

Finally, we generate the three-channel mask by
\begin{equation}
    \boldsymbol{M}_c = Sgn(\boldsymbol{S}_{rank} \odot \boldsymbol{M}_b), 
\end{equation} where $Sgn(\cdot)$ represents a function generating the final three-channel mask based on the signs of the values in $\boldsymbol{S}_{rank} \odot \boldsymbol{M}_b$: 0 results in [0, 0, 1] (blue), positive values result in [0, 1, 0] (green), and negative values result in [1, 0, 0] (red).
\begin{table*}[t!]
    \centering
	\caption{Discrimination performance comparisons on the synthetic dataset.} 
	\label{table:Synthetic3} 
\begin{tabular}{ccccccccccccc}
\hline
\multirow{2}{*}{Methods} & \multicolumn{3}{c}{Background} & \multicolumn{1}{c}{} & \multicolumn{3}{c}{Source}& \multicolumn{1}{c}{} & \multicolumn{3}{c}{Target} \\ \cline{2-4} \cline{6-8} \cline{10-12} 
                         & Precision    & Recall    & F1 &   & Precision   & Recall   & F1 &   & Precision  & Recall  & F1  \\ \hline
BusterNet\cite{ferrari_busternet_2018} & 0.935 & 0.995 & 0.963 & & 0.098 & 0.048 & 0.053 & & 0.226 & 0.087 & 0.108\\
BusterNet (retrain) &0.972  &0.982  &0.977  & &0.204  &0.190  &0.162  & &0.746  &0.762  &0.735 \\
DOA-GAN\cite{islam_doa-gan_2020} & 0.931 & 0.997 & 0.962 & & 0.028 & 0.009 & 0.011 & & 0.187 & 0.065 & 0.081\\
DOA-GAN (retrain) & 0.974 & 0.998 & 0.986 & & 0.604 & 0.210 & 0.289 & & 0.906 & 0.924 & 0.911\\
Serial Network$^{*}$\cite{chen_serial_2020} & 0.938 & 0.981 & 0.958 & & 0.088 & 0.046 & 0.092 & & 0.158 & 0.077 & 0.085\\
Serial Network (retrain)$^{*}$ & 0.966 & 0.990 & 0.980 & & 0.055 & 0.034 & 0.039 & & 0.838 & 0.488 & 0.584\\
D2PRL & \textbf{0.991} & \textbf{0.998} & \textbf{0.994} & & \textbf{0.906} & \textbf{0.773} & \textbf{0.817} & & \textbf{0.914} & \textbf{0.935} & \textbf{0.917}\\
\hline
\end{tabular}
\end{table*}

\begin{table}[t!]
	\centering
	\caption{Single-channel localization performance comparisons on the synthetic dataset.} 
	\label{table:Synthetic1}  
	\begin{tabular}{c|ccc}
		\hline
		Methods & Precision & Recall & F1 \\
		\hline
        BusterNet\cite{ferrari_busternet_2018} & 0.270 & 0.080 & 0.107\\
        BusterNet (retrain) &0.671  &0.504  &0.544 \\
		DOA-GAN\cite{islam_doa-gan_2020} & 0.656 & 0.269 & 0.329\\
		DOA-GAN (retrain) & 0.911 & 0.587 & 0.694\\
		Serial Network\cite{chen_serial_2020} & 0.248 & 0.127 & 0.144\\
		Serial Network (retrain) & 0.856 & 0.498 & 0.583\\
		D2PRL & \textbf{0.947} & \textbf{0.767} & \textbf{0.832}\\
		\hline
	\end{tabular}
\end{table}

\subsection{Loss Functions}\label{sec:loss}
In this work, we propose training the network using a fused loss combined with a similarity localization loss DFM $L_{DFM}$, a target localization loss MRD $L_{MRD}$ and the discrimination loss $L_{dis}$
\begin{equation}
		\begin{aligned}
		L = L_{DFM}+L_{MRD}+L_{dis}.
		\end{aligned}
\end{equation}

The similarity localization loss $L_{DFM}$ is to ensure the accuracy of the predicted single-channel mask $\boldsymbol{M}$, which is essentially a binary classification problem for each pixel. In our work, we directly adopt the dice loss
\begin{equation}
		\begin{aligned}
		L_{DFM} = 1 -\frac{2\times\sum_{i,j}\boldsymbol{M}^{gt}(i,j)\boldsymbol{M}(i,j)}{\sum_{i,j}\boldsymbol{M}^{gt}(i,j)+\sum_{i,j}\boldsymbol{M}(i,j)},
		\end{aligned}
\end{equation}
where ${\boldsymbol{M}}$ is the predicted single-channel mask and $\boldsymbol{M}^{gt}$ is the ground-truth single-channel mask for source and target regions. 

The target localization loss $L_{MRD}$ is to ensure the accuracy of the predicted target regions mask $\boldsymbol{M}_{t}$. Similarly, we directly adopt the dice loss to compute $L_{MRD}$, i.e.,
\begin{equation}
		\begin{aligned}
		L_{MRD} = 1 -\frac{2\times\sum_{i,j}\boldsymbol{M}_t^{gt}(i,j)\boldsymbol{M}_{t}(i,j)}{\sum_{i,j}\boldsymbol{M}_t^{gt}(i,j)+\sum_{i,j}\boldsymbol{M}_{t}(i,j)},
		\end{aligned}
\end{equation}
where $\boldsymbol{M}_t^{gt}$ represents the ground-truth mask of the target region.

The discrimination loss $L_{dis}$ is to ensure that $\boldsymbol{S}_{rank}(i,j)$ of the source region is greater than 0, and $\boldsymbol{S}_{rank}(i,j)$ of the target region is less than 0. To achieve this, we introduce a weight matrix $\boldsymbol{W}$, whose elements are 
\begin{equation}
		\boldsymbol{W}(i,j) = 
        \begin{cases}
        -1       &(i,j)\in{source\ regions}\\
        1   &(i,j)\in{target\ regions}\\
        0   &otherwise
        \end{cases},
\end{equation} Note that $\boldsymbol{W}$ is only related to the ground-truth three-channel mask $\boldsymbol{M}_c^{gt}$.

We further define
\begin{equation}
	\boldsymbol{\tilde{S}}_{rank}(i,j)=\boldsymbol{S}_{rank}(i,j)\boldsymbol{W}(i,j).
\end{equation} It should be noted that $\boldsymbol{\tilde{S}}_{rank}(i,j)\leq 0$ means that the pixel $(i,j)$ is from the background or it is correctly discriminated by our framework. $\boldsymbol{\tilde{S}}_{rank}(i,j) > 0$ indicates that our framework either incorrectly classifies target pixels as source pixels, or it erroneously classifies source pixels as target pixels. As a result, we need only put a penalty when $\boldsymbol{\tilde{S}}_{rank}(i,j) > 0$. In our work, we further introduce a margin penalty to improve the robustness, and the final discrimination loss is defined as
\begin{equation} \label{eq:Ldis}
		L_{dis} = \sum_{i,j}
    \begin{cases}
        0       &\boldsymbol{\tilde{S}}_{rank}(i,j) \leq \tau \ or \ \boldsymbol{M}^{gt}(i,j)=0\\
        \boldsymbol{\tilde{S}}_{rank}(i,j)-\tau   &otherwise
    \end{cases}
\end{equation} 
where $\tau$ is a small negative constant, empirically set to $-0.05$ in our experiments. By resorting to the proposed pairwise rank learning method, our framework is enforced to learn subtle clues to correctly discriminate the matched points. It is worth noting that all the components of our framework are differentiable, which allows us to train our framework in an end-to-end manner.

\section{Experiments}\label{sec:Experiment Result}
In this section, we evaluate the effectiveness of our scheme under a series of experimental settings.

\subsection{Implementation Details}
We implement the proposed model D2PRL  using the PyTorch framework. Input images are initially resized to a standardized size of 448$\times$448. Convolution and BatchNorm layers are initialized using kaiming initialization. During the initial training phase, only $L_{DFM}$ and $L_{MRD}$ are utilized to train the localization component, employing the Adam optimizer with a learning rate of 1e-3. After three epochs, we use the fused loss $L$ to train the whole framework with the learning rate adjusted to 1e-4. All experiments in this paper are implemented on one NVIDIA RTX3090 GPU with an Intel(R) Xeon(R) Silver 4210 CPU.

\subsection{Dataset}
\subsubsection{Training dataset}
In our study, we utilize images from MS COCO 2014 \cite{lin2014microsoft} to create a synthetic training dataset. Each image is resized to dimensions of 1024$\times$1024, and random polygonal or annotated object regions are selected and inserted into other parts of the image. To enhance the model's generalization across diverse scenes, we address both cases where copy-move manipulation occurs within objects and backgrounds. Notably, background copy-move forgery is a common real-world practice to use background areas for concealing specific objects of interest. In our dataset, we maintain a ratio of 8:2 between random polygonal and annotated object regions for good generalization capability. It's worth mentioning that all copied snippets are subject to random rotation [-180°,180°] and scaling [0.5,2] attacks. Adhering to this approach, we generate a total of 99,353 training images. Prior to network training, these training images are subject to random JPEG compression and Gaussian noise addition to ensure robust training.
\subsubsection{Testing dataset}
\begin{itemize}
    \item Synthetic dataset: this dataset has 4,000 testing images, which are generated in the same way as the training set.
    \item CASIA CMFD dataset\cite{dong2013casia}: this dataset comprises 1309 copy-move forgery images, all of which are manually created to include rotation and scaling transformation attacks.
    \item CoMoFoD dataset\cite{tralic_comofod_2013}: this dataset consists of 200 base forgery images, each subjected to 24 different post-processing/attack techniques, resulting in a total of 4,800 forgery images.
    \item {CMH\cite{silva_going_2015}}: this dataset contains 108 forgery images featuring rotation and scaling transformations. To assess our model's discrimination performance, we manually annotate the source and target for each image.
\end{itemize}

It should be emphasized that we exclusively train our framework on the training set and directly apply it to other testing datasets without fine-tuning.

\begin{figure}[t!]
	\centering
	\includegraphics[width=1\linewidth]{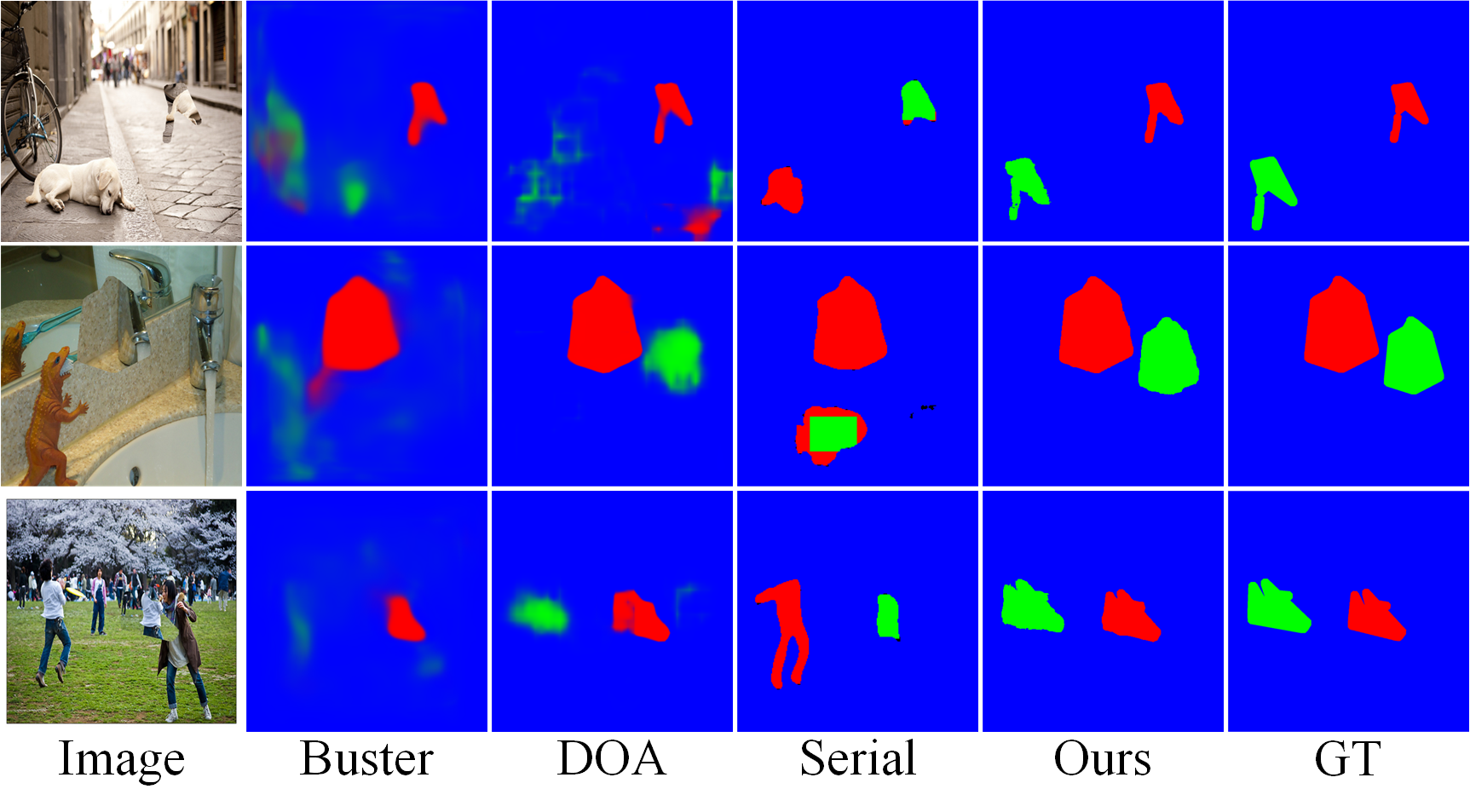}
	\centering
    \caption{Visual comparison on the Synthetic dataset.}
    \label{fig:testn_visual}
\end{figure}

\subsection{Comparison with State-of-the-art Methods}
In this section, we compare our method D2PRL to state-of-the-art approaches, evaluating performance through pixel-level precision, recall, and F1 scores. Please note that these scores are computed individually for each image and then averaged over all images in the dataset.
\subsubsection{Experimental results on Synthetic dataset}
In this subsection, we assess our algorithm's performance on the synthetic dataset and compare it with three deep learning methods: Busternet\cite{ferrari_busternet_2018}, Serial Network\cite{chen_serial_2020}, DOA-GAN\cite{islam_doa-gan_2020}. These methods were selected for their ability to differentiate source and target regions and their availability of public source code. For fairness, Table \ref{table:Synthetic3} provides the performance of the pre-trained models of all three methods. Additionally, we also present the performance of these models after retraining on our synthetic dataset, where the image resolution is $448\times 448$, consistent with our method. We mark the serial network with an asterisk $*$, as it can only disambiguate the source and target regions for single-channel masks that have exactly two disconnected regions. For masks where it cannot complete the disambiguation, we randomly assign each detected pixel to either the source or target region as output, and then calculate average performance across all images.


Table \ref{table:Synthetic3} demonstrates that our model achieves the highest performance in all cases. particularly demonstrating remarkable gains in source region detection. We can observe that 
before retraining, Busternet, Serial Network and DOA-GAN perform poorly on both source region and target region detection. One possible reason is that the synthetic dataset contains many non-object copy-move images, while the above pre-trained models overfitted to objects are ineffective in revealing such copy-move traces. 
After retraining, Busternet, Serial Network and DOA-GAN consistently demonstrate good performance in target region detection but still struggled in source region detection. This is because these methods effectively learned the forged traces left by pasting and post-processing methods in the target regions.  However, their reliance on high-level semantic features made it difficult for them to establish a matching relationship between non-object sources and target regions, thus making them ineffective in detecting the source regions. In contrast, our model relies on low-level semantic features that focus on texture information rather than object-level features, allowing it to establish accurate matching relationships and uncover the source regions. 
In addition to the three-channel performance, Table \ref{table:Synthetic1} also presents the single-channel results of various algorithms. We can still observe that our D2PRL method consistently outperforms the competing approaches.


Fig. \ref{fig:testn_visual} presents some visual comparisons among different methods, where blue depicts the detected background, green depicts the detected source regions, and red depicts the detected target regions. Notably, the three other methods only perform relatively well in detecting target regions but struggle to accurately identify source regions. The serial network, for instance, solely detects the person's outline in the third image, demonstrating overfitting to high-level semantic object features. In contrast, our model's masks exhibit notably higher accuracy and granularity, surpassing the performance of other approaches.

\begin{table}[t!]
	\centering
	\caption{Single-channel localization results on CASIA CMFD dataset. Results marked with "$\dagger$"  are from \cite{zhong_end--end_2020}.} 
	\label{table:casia}  
	\begin{tabular}{c|ccc}
		\hline
		Methods & Precision & Recall & F1 \\ \hline
		Zernike$^{\dagger}$\cite{ryu_rotation_2013} & 0.227 & 0.134 & 0.164\\
		PCET$^{\dagger}$\cite{emam_pcet_2016} & 0.311 & 0.294 & 0.302\\
		MSA$^{\dagger}$\cite{pan_region_2010} & 0.557 & 0.543 & 0.555\\
		SURF$^{\dagger}$\cite{shivakumar2011detection} & 0.359 & 0.340 & 0.349\\
		OverSeg$^{\dagger}$\cite{pun_image_2015} & 0.430 & 0.367 & 0.396\\
		HFPM$^{\dagger}$\cite{li_fast_2019} & 0.578 & 0.659 & 0.616\\
		PM$^{\dagger}$\cite{cozzolino_efficient_2015} & 0.473 & 0.495 & 0.484\\
		ECSS$^{\dagger}$\cite{bi2018fast} & 0.420 & 0.459 & 0.439\\
        SSG\cite{10007894} & 0.827 & \textbf{0.817} & 0.796\\
		BusterNet\cite{ferrari_busternet_2018} & 0.557 & 0.438 & 0.456\\
		DOA-GAN\cite{islam_doa-gan_2020} & 0.547 & 0.397 & 0.414\\
		Serial Network\cite{chen_serial_2020} & 0.531 & 0.498 & 0.477\\
		DFIC\cite{zhong_end--end_2020} & 0.709 & 0.589 & 0.643\\
        DMPS\cite{liu2021two}& 0.636 & 0.475 & 0.492\\
        UCM-Net-s\cite{weng2023ucm}& 0.590 & 0.630 & 0.580\\
		D2PRL & \textbf{0.875} & 0.784 & \textbf{0.806}\\
		\hline
	\end{tabular}
\end{table}

\begin{table*}[t!]
    \centering
	\caption{Three-channel localization results on CASIA CMFD dataset.} 
	\label{table:casia2} 
\begin{tabular}{cccccccccccc}
\hline
\multirow{2}{*}{Methods} & \multicolumn{3}{c}{Background} & \multicolumn{1}{c}{} & \multicolumn{3}{c}{Source} & \multicolumn{1}{c}{} & \multicolumn{3}{c}{Target} \\ \cline{2-4} \cline{6-8} \cline{10-12}
                         & Precision    & Recall    & F1 &   & Precision   & Recall   & F1 &    & Precision  & Recall  & F1  \\ \hline
BusterNet\cite{ferrari_busternet_2018} & 0.953 & 0.985 & 0.967 & & 0.297 & 0.344 & 0.283 & & 0.211 & 0.089 & 0.103\\
DOA-GAN\cite{islam_doa-gan_2020} & 0.9072 & 0.998 & 0.947 & & 0.148 & 0.068 & 0.077 & & 0.121 & 0.055 & 0.063\\
Serial Network$^{*}$\cite{chen_serial_2020} & 0.960 & 0.977 & 0.967 & & 0.285 & 0.236 & 0.234 & & 0.252 & 0.307 & 0.251\\
D2PRL & \textbf{0.9762} & \textbf{0.992} & \textbf{0.983} & & \textbf{0.541} & \textbf{0.585} & \textbf{0.543} & & \textbf{0.571} & \textbf{0.557} & \textbf{0.545}\\
\hline
\end{tabular}
\end{table*}

\subsubsection{Experimental results on CASIA CMFD dataset}
The CASIA CMFD dataset is much more challenging because all of the copy-move operations are implemented manually so that the target regions are blended very well with the background and a certain number of copy-move manipulations occur in the background. 

Since most existing algorithms can only detect single-channel forgery maps, we first evaluate the performance of different algorithms in terms of similarity localization, including nine conventional algorithms based on hand-crafted features (Zernike\cite{ryu_rotation_2013}, PCET\cite{emam_pcet_2016}, MSA\cite{pan_region_2010}, SURF\cite{shivakumar2011detection}, OverSeg\cite{pun_image_2015}, HFPM\cite{li_fast_2019}, PM\cite{cozzolino_efficient_2015}, ECSS\cite{bi2018fast},SSG\cite{10007894}) and six deep learning-based algorithms (BusterNet\cite{ferrari_busternet_2018},DOA-GAN\cite{islam_doa-gan_2020}, Serial Network\cite{chen_serial_2020}, DFIC\cite{zhong_end--end_2020}, DMPS\cite{liu2021two}, UCM-Net-s\cite{weng2023ucm}). 
Note that for BusterNet, DOA-GAN, and Serial Network, we found that their methods trained on our synthetic dataset exhibited very poor generalization on other datasets. Therefore, we used their pre-trained models to evaluate their performance on other datasets.
Table \ref{table:casia} summarizes the localization results, clearly highlighting our method's superiority over both conventional and deep learning approaches. Specifically, our model achieves an F1 score 16\% higher than the second-best deep learning model DFIC. 
The closest performance to our method is achieved by the conventional keypoint-based approach SSG. However, SSG requires resizing the input image's long side to 3000 pixels while maintaining the original aspect ratio, whereas our model operates at a resolution of 448$\times$448 pixels. Furthermore, as one of the conventional methods, SSG lacks the capability to differentiate between source and target regions.

 


We further compare our method to Busternet \cite{ferrari_busternet_2018}, Serial Network\cite{chen_serial_2020}, DOA-GAN\cite{islam_doa-gan_2020} in terms of the three-channel performance. The results are presented in Table \ref{table:casia2}. 
The asterisk (*) indicates that the Serial Network was successful in generating three-channel masks for only 747 images. For the remaining images, we randomly assign each detected pixel to either the source or target region as output.
It can be observed that our model significantly outperforms the other three methods. We attribute this improvement to the utilization of point-to-point low-level feature matching and ranking learning, which enable our method to avoid overfitting objects in the training images and force the model to learn subtle cues for distinguishing source and target regions. It should be emphasized that such merits cannot simply be obtained by the existing CNN-based approaches. 

\begin{figure}[t!]
	\centering
	\includegraphics[width=1\linewidth]{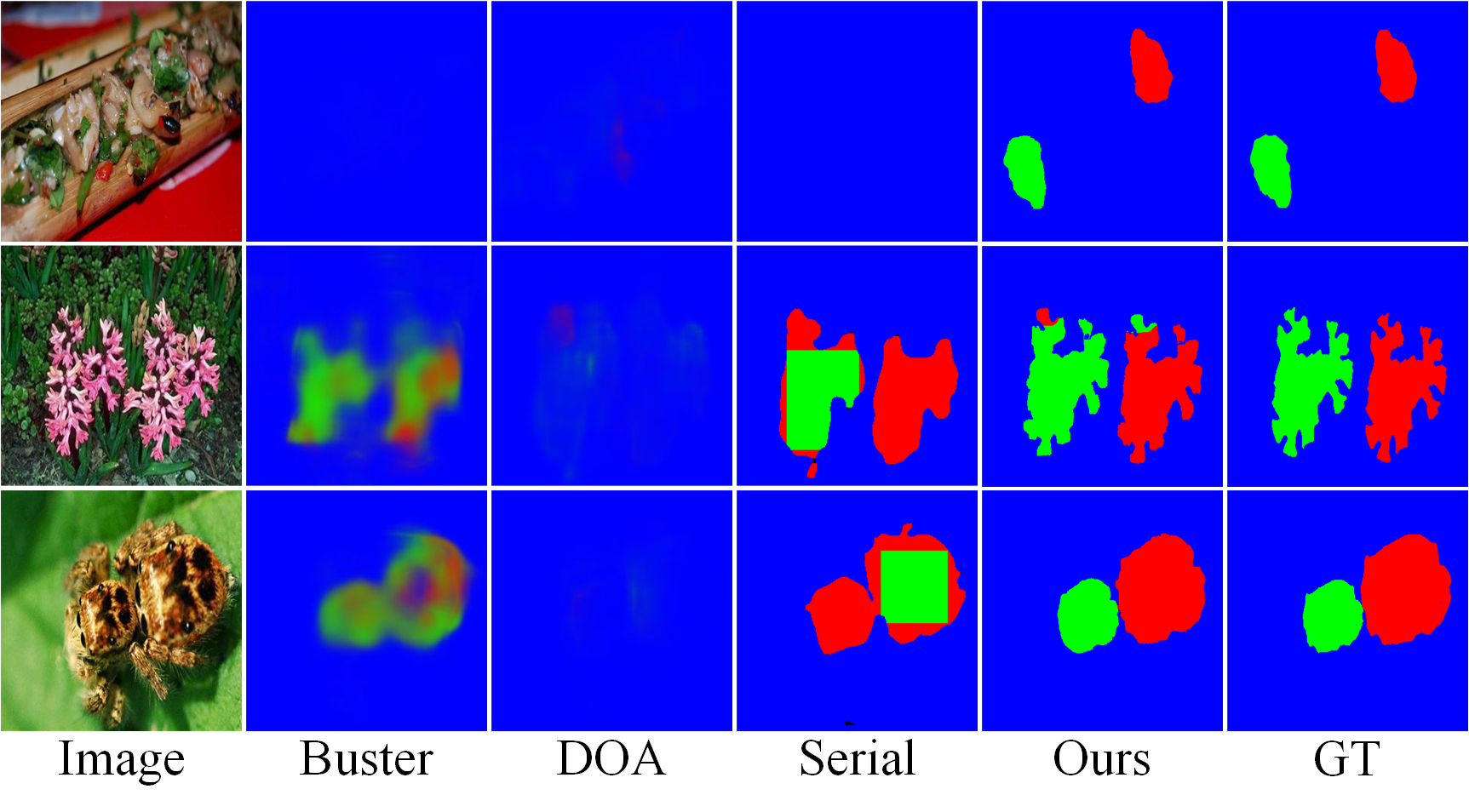}
	\centering
	\caption{Visual comparison on the CASIA CMFD dataset.}
	\label{fig:casia_v}
\end{figure}

Fig. \ref{fig:casia_v} illustrates the visual results obtained from CASIA CMFD. 
The first image showcases the effectiveness of our method in detecting non-object copy-move instances with rotations, unsurprisingly, only our method can detect the copy-move regions with source-target discrimination. The second image illustrates our method generates masks with finer edges due to the high-resolution feature maps and point-to-point matching. In the third image, where the crab's shell has been copied and significantly resized before pasting, we can observe that Busternet, Serial Network, and our method are all capable of detecting the forged regions. However, only our method successfully distinguishes between the source and target regions.

\subsubsection{Experimental results on CMH dataset}
In this subsection, we further demonstrate the generalizability of our method on the CMH dataset. Firstly, we compare the single-channel localization performance with conventional algorithms including PM\cite{cozzolino_efficient_2015}, Iteration\cite{zandi2016iterative}, HFPM\cite{li_fast_2019}, CMI\cite{niu2021fast}, and SSG\cite{10007894}, as well as deep models such as BusterNet\cite{ferrari_busternet_2018}, DOA-GAN\cite{islam_doa-gan_2020}, and Serial Network\cite{chen_serial_2020}. The localization performance results are listed in Table \ref{table:CMH_1c}.
We can see that our method also generalizes very well to the CMH dataset, the obtains the best F1 score among all the competing algorithms. It's noticeable that conventional methods achieve good performance on the CMH dataset due to its relatively monotonic forgery nature and simpler image backgrounds compared to the CASIA CMFD dataset. However, previous deep learning algorithms like BusterNet, DOA-GAN, and Serial Network show poorer performance. We conjecture that this is due to the prevalence of objects not present in their training datasets, and a significant portion of these objects are not involved in copy-move forgeries.

\begin{figure}[t!]
	\centering
	\includegraphics[width=1\linewidth]{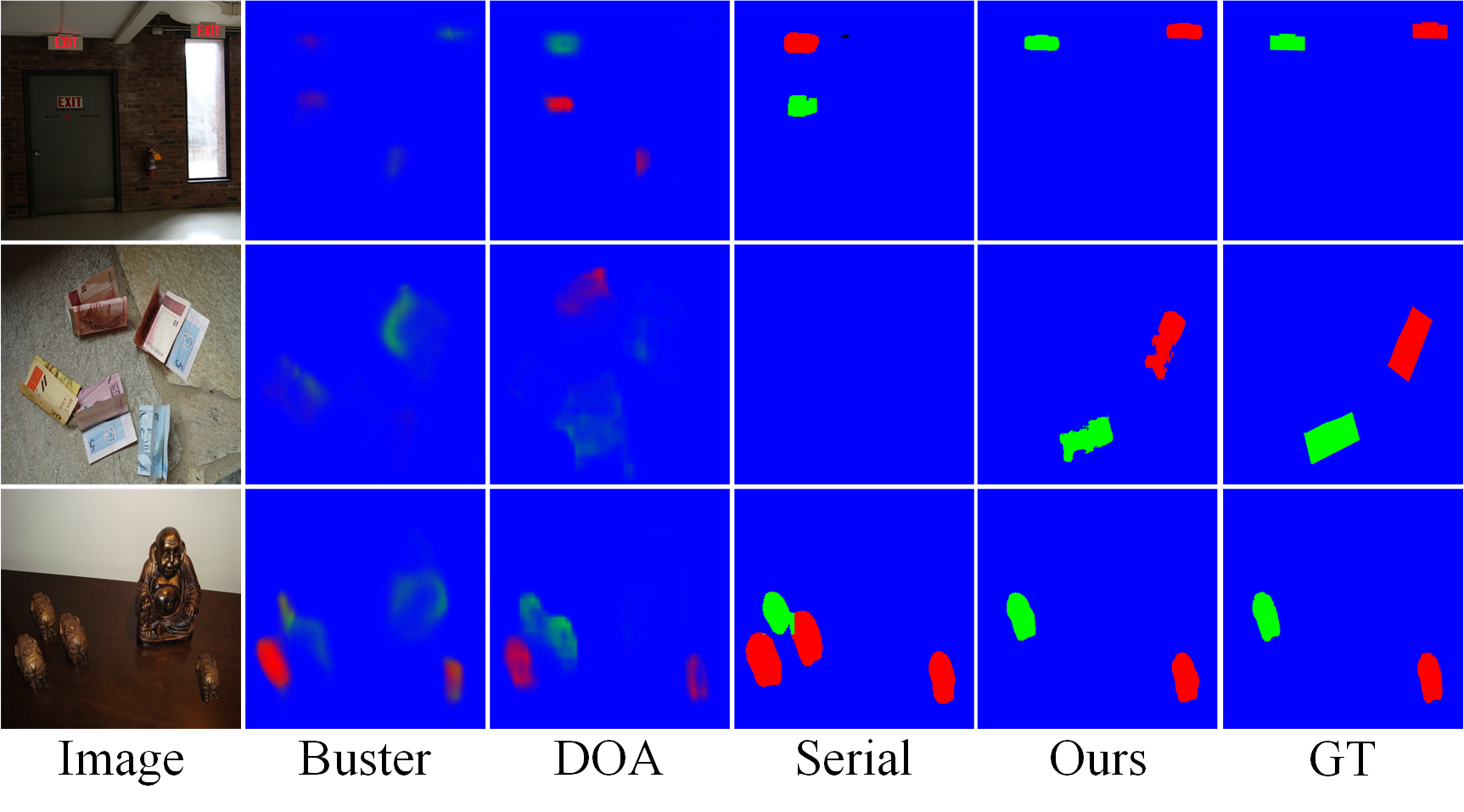}
	\centering
    \caption{Visual comparison of discrimination results on CMH dataset.}
    \label{fig:CMH_v}
\end{figure}

Next, we compare the three-channel localization performance with BusterNet, DOA-GAN and Serial Network, and the results are summarized in Table \ref{table:CMH_3c}. The asterisk (*) indicates that the Serial Network was successful in generating three-channel masks for only 55 images. For the remaining images, we randomly assign each detected pixel to either the source or target region as output. Our method, D2PRL, significantly outperforms the competing deep models in discriminating the source and target regions. Fig. \ref{fig:CMH_v} showcases some visual results on CMH dataset. We can observe that BusterNet, DOA-GAN, and Serial Network tend to misclassify certain genuine objects as copy-move regions, and they also struggle in effectively distinguishing between the source and target regions. In contrast, our method consistently exhibits superior performance, even when faced with severe rotations.

\begin{table}[t!]
	\centering
	\caption{Single-channel localization results on CMH dataset. Results marked with "$\dagger$"  are from \cite{10007894}.} 
	\label{table:CMH_1c}  
	\begin{tabular}{c|ccc}
		\hline
		Methods & Precision & Recall & F1 \\ \hline
		PM$^{\dagger}$\cite{cozzolino_efficient_2015} & 0.830 & 0.790 & 0.801\\
        Iteration$^{\dagger}$\cite{zandi2016iterative} & 0.550 & 0.653 & 0.583\\
        HFPM$^{\dagger}$\cite{li_fast_2019} & 0.853 & 0.720 & 0.764\\
        CMI$^{\dagger}$\cite{niu2021fast} & 0.798 & \textbf{0.885} & 0.803\\            
		SSG\cite{10007894} & 0.844 & 0.814 & 0.822\\
		BusterNet\cite{ferrari_busternet_2018} & 0.330 & 0.420 & 0.336\\
		DOA-GAN\cite{islam_doa-gan_2020} & 0.530 & 0.340 & 0.364\\
		Serial Network\cite{chen_serial_2020} & 0.412 & 0.392 & 0.369\\
		D2PRL & \textbf{0.911} & 0.811 & \textbf{0.846}\\
		\hline
	\end{tabular}
\end{table}

\begin{table*}[t!]
    \centering
	\caption{Three-channel localization results on CMH dataset.} 
	\label{table:CMH_3c} 
\begin{tabular}{cccccccccccc}
\hline
\multirow{2}{*}{Methods} & \multicolumn{3}{c}{Background} & \multicolumn{1}{c}{} & \multicolumn{3}{c}{Source} & \multicolumn{1}{c}{} & \multicolumn{3}{c}{Target} \\ \cline{2-4} \cline{6-8} \cline{10-12}
                         & Precision    & Recall    & F1 &   & Precision   & Recall   & F1 &    & Precision  & Recall  & F1  \\ \hline
BusterNet\cite{ferrari_busternet_2018} & 0.974 & 0.993 & 0.983 & & 0.200 & 0.173 & 0.176 & & 0.210 & 0.157 & 0.147\\
DOA-GAN\cite{islam_doa-gan_2020} & 0.963 & 0.995 & 0.979 & & 0.058 & 0.032 & 0.037 & & 0.046 & 0.031 & 0.036\\
Serial Network$^{*}$\cite{chen_serial_2020} & 0.977 & 0.982 & 0.979 & & 0.216 & 0.182 & 0.184 & & 0.192 & 0.198 & 0.173\\
D2PRL & \textbf{0.994} & \textbf{0.998} & \textbf{0.996} & & \textbf{0.632} & \textbf{0.609} & \textbf{0.611} & & \textbf{0.660} & \textbf{0.599} & \textbf{0.615}\\
\hline
\end{tabular}
\end{table*}

\begin{table}[t!]
	\centering
	\caption{Single-channel localization results on CoMofoD dataset without attacks. Results marked with ``$\dagger$"  are from \cite{chen_serial_2020}.
 } 
	\label{table:comofod}  
	\begin{tabular}{c|ccc}
		\hline
		Methods & Precision & Recall & F1 \\ \hline
		OverSeg$^{\dagger}$\cite{pun_image_2015} & 0.349 & 0.214 & 0.222\\
		PM$^{\dagger}$\cite{cozzolino_efficient_2015} & 0.476 & 0.399 & 0.418\\
        HFPM\cite{li_fast_2019} & 0.428 & 0.424 & 0.426\\
        SSG\cite{10007894} & 0.572 & 0.536 & 0.507\\
        BusterNet\cite{ferrari_busternet_2018} & {0.612} & {0.454} & {0.508}\\
		DOA-GAN\cite{islam_doa-gan_2020} & 0.552 & 0.416 & 0.420\\
        Serial Network\cite{chen_serial_2020} & 0.531 & 0.498 & 0.477\\
		DFIC\cite{zhong_end--end_2020} & 0.461 & 0.422 & 0.441\\
        DMPS\cite{liu2021two} & 0.618 & 0.544 & 0.517\\
        UCM-Net-s\cite{weng2023ucm} & 0.620 &0.600 &0.580\\
		D2PRL & \textbf{0.763} & \textbf{0.818} & \textbf{0.758}\\
		\hline
	\end{tabular}
\end{table}
\begin{figure*}[t!]
	\centering
	\includegraphics[width=0.74\linewidth]{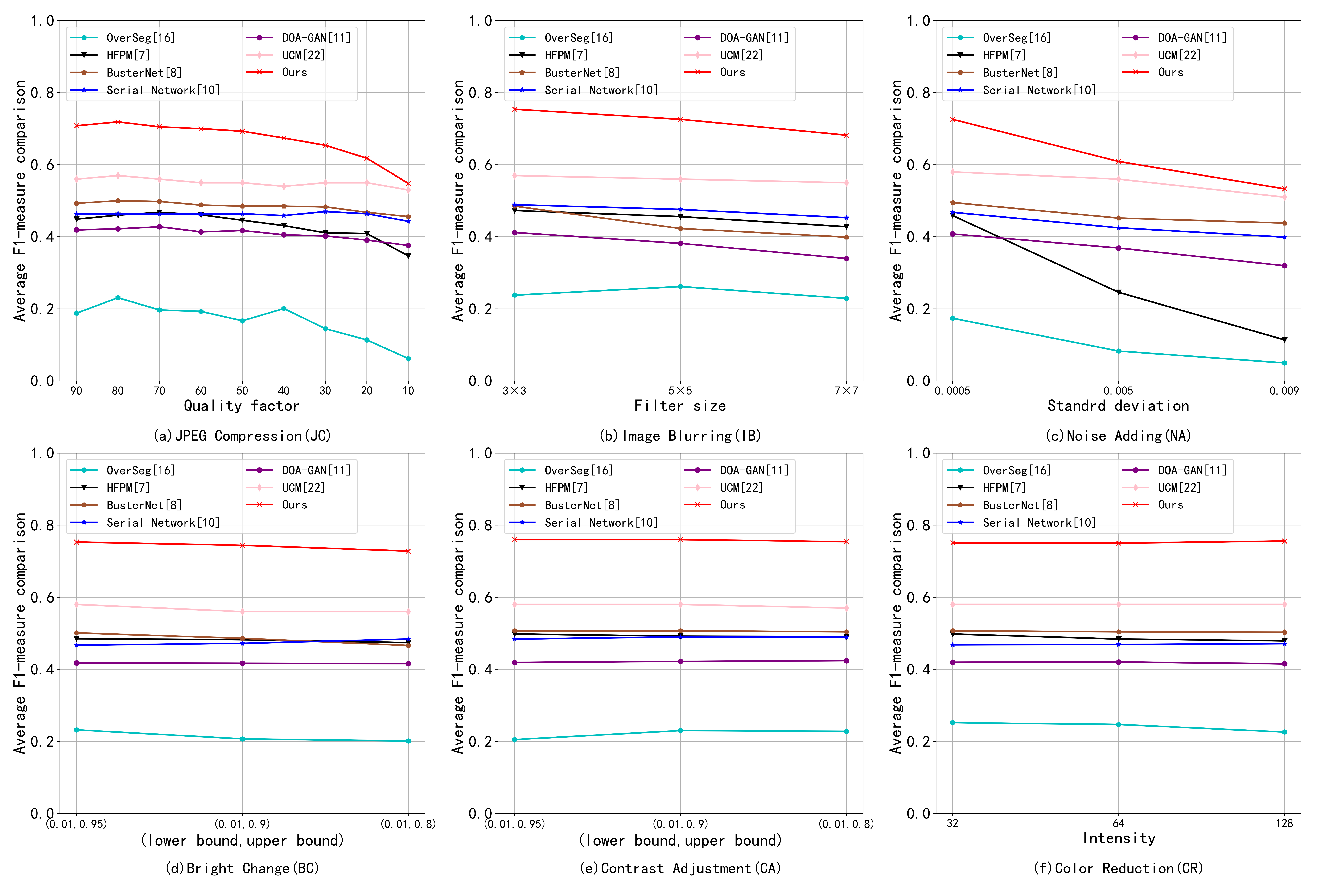}
	\centering
	\caption{F1 comparisons on CoMoFoD dataset under different attacks.}
	\label{fig:comoR}
\end{figure*}
\begin{figure}
	\centering
	\includegraphics[width=1\linewidth]{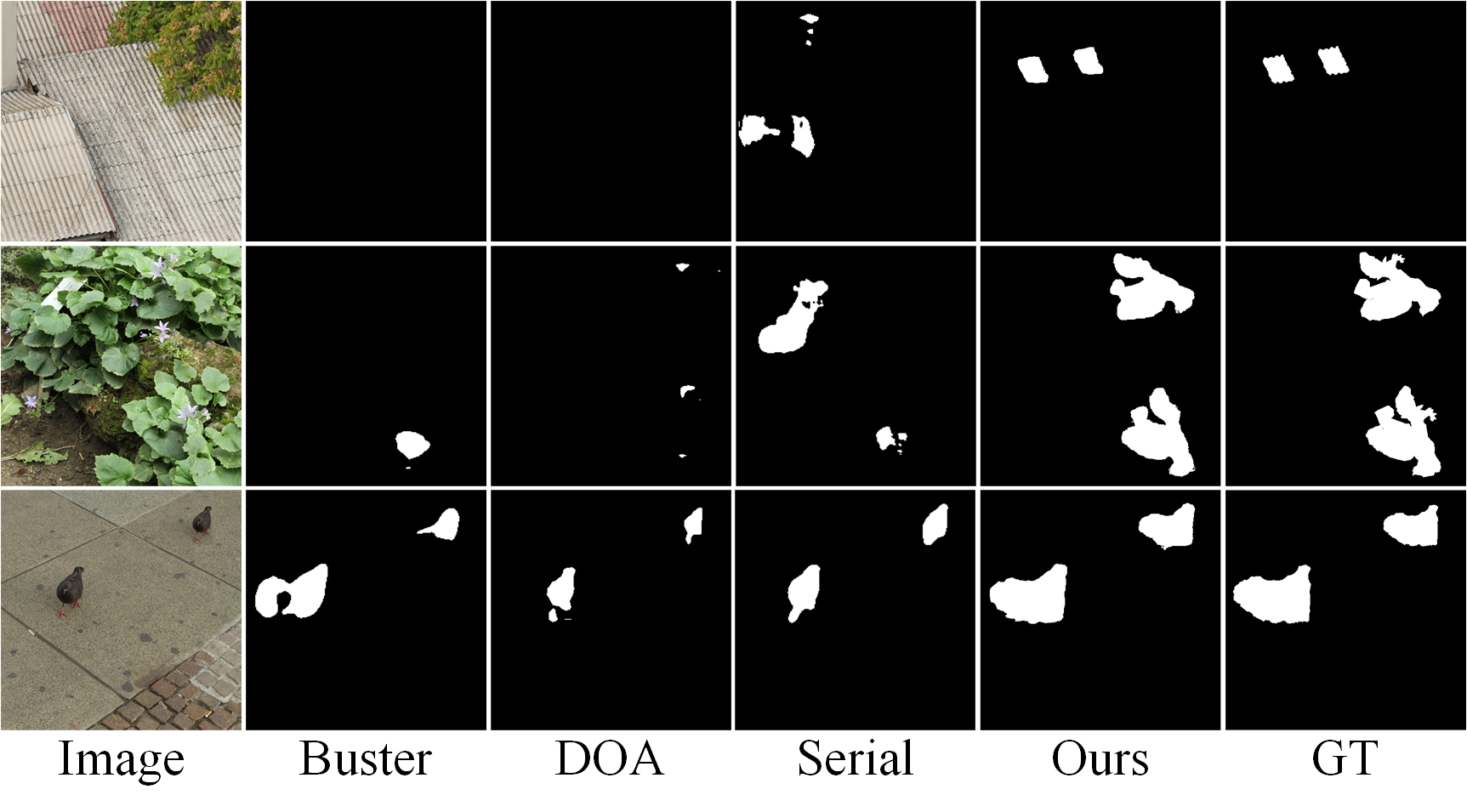}
	\centering
    \caption{Visual comparison of localization results on the CoMoFoD dataset.}
    \label{fig:comofod_visual}
\end{figure}
\begin{table*}[t!]
 \centering
 \caption{Discrimination performance comparison on detectable images for DisTool.} 
	\label{table:DisTool_1}
\begin{tabular}{ccccccclccc}
\hline
\multirow{2}{*}{Dataset} && \multirow{2}{*}{Methods} && \multicolumn{3}{c}{Source} &  & \multicolumn{3}{c}{Target} \\ \cline{5-7} \cline{9-11} 
                         &&                          && Precision & recall & F1    & & Precision & recall & F1    \\ \hline
\multirow{2}{*}{SYN}     && DisTool                  && 0.840     & 0.700  & 0.748 & & 0.827     & 0.834  & 0.825 \\
                         && D2PRL                    && \textbf{0.952}     & \textbf{0.822}  & \textbf{0.866} & & \textbf{0.965}     & \textbf{0.873}  & \textbf{0.908} \\ \hline
\multirow{2}{*}{CASIA}   && DisTool                 && 0.617     & 0.621  & 0.602 & & 0.622     & 0.605  & 0.579 \\
                         && D2PRL                    && \textbf{0.636}     & \textbf{0.664}  & \textbf{0.630} & & \textbf{0.639}     & \textbf{0.634}  & \textbf{0.619} \\ \hline
\multirow{2}{*}{CMH}     && DisTool                 && 0.549     & 0.534  & 0.538 & & 0.552     & 0.527  & 0.534 \\
                         && D2PRL                    && \textbf{0.597}     & \textbf{0.628}  & \textbf{0.601} & & \textbf{0.613}     & \textbf{0.562}  & \textbf{0.575} \\ \hline
\end{tabular}
\end{table*}
\subsubsection{Experimental results on CoMoFoD dataset}
The CoMoFoD dataset presents a significant challenge for detection algorithms due to its abundance of smooth and small copy-move regions.  As the Comofod dataset does not provide ground-truth masks to differentiate between source and target regions, our experiments only compare single-channel performance.  Table \ref{table:comofod} provides the performance comparison of our method with other state-of-the-art techniques on 200 base images.  It is evident that our method outperforms others significantly, as the utilization of low-level semantic features and dense-field matching effectively detects smooth and small regions. 

To further validate the robustness of our method against different types of attacks, we divide the images from the CoMoFoD dataset into six groups based on the type of attack: JPEG compression (JC), Image blurring (IB), Noise adding (NA), Bright Change (BC), Contrast adjustment (CA), and Color reduction (CR). The comparison of F1 scores with state-of-the-art methods is shown in Fig. \ref{fig:comoR}.  Remarkably, our method achieves much better results across all six types of attacks compared to other models, even when faced with severe JPEG compression and noise addition.  This further demonstrates the strong robustness of our method against various attacks. Fig. \ref{fig:comofod_visual} presents visualization results on the CoMoFoD dataset, which demonstrate that our method can successfully detect small and smooth copy-move regions.

\begin{table}[t!]
	\centering
	\caption{Ablation experiment with F1-scores on Synthetic dataset.}
	\label{table:ablation2}  
 \scalebox{0.85}{
	\begin{tabular}{ccccc|ccc}
		\hline
		CNN & ZM & DLF & CS-PM & SE-U-Net &  Background & Source & Target \\ \hline
		  & \checkmark & \checkmark & \checkmark & \checkmark & 0.984 & 0.677 & 0.737 \\
            \checkmark &  & \checkmark & \checkmark & \checkmark & 0.988 & 0.717 & 0.735 \\
		\checkmark & \checkmark &  & \checkmark & \checkmark & 0.988 & 0.749 & 0.814 \\ 
		\checkmark & \checkmark & \checkmark &  & \checkmark & 0.979 & 0.577 & 0.627 \\ 
            \checkmark & \checkmark & \checkmark & \checkmark & &0.991  &0.817 &0.866 \\
            \checkmark & \checkmark & \checkmark & \checkmark & \checkmark & \textbf{0.994} & \textbf{0.817} & \textbf{0.917} \\
		\hline
        \end{tabular}}
\end{table}
\begin{figure}[t!]
	\centering
	\includegraphics[width=1\linewidth]{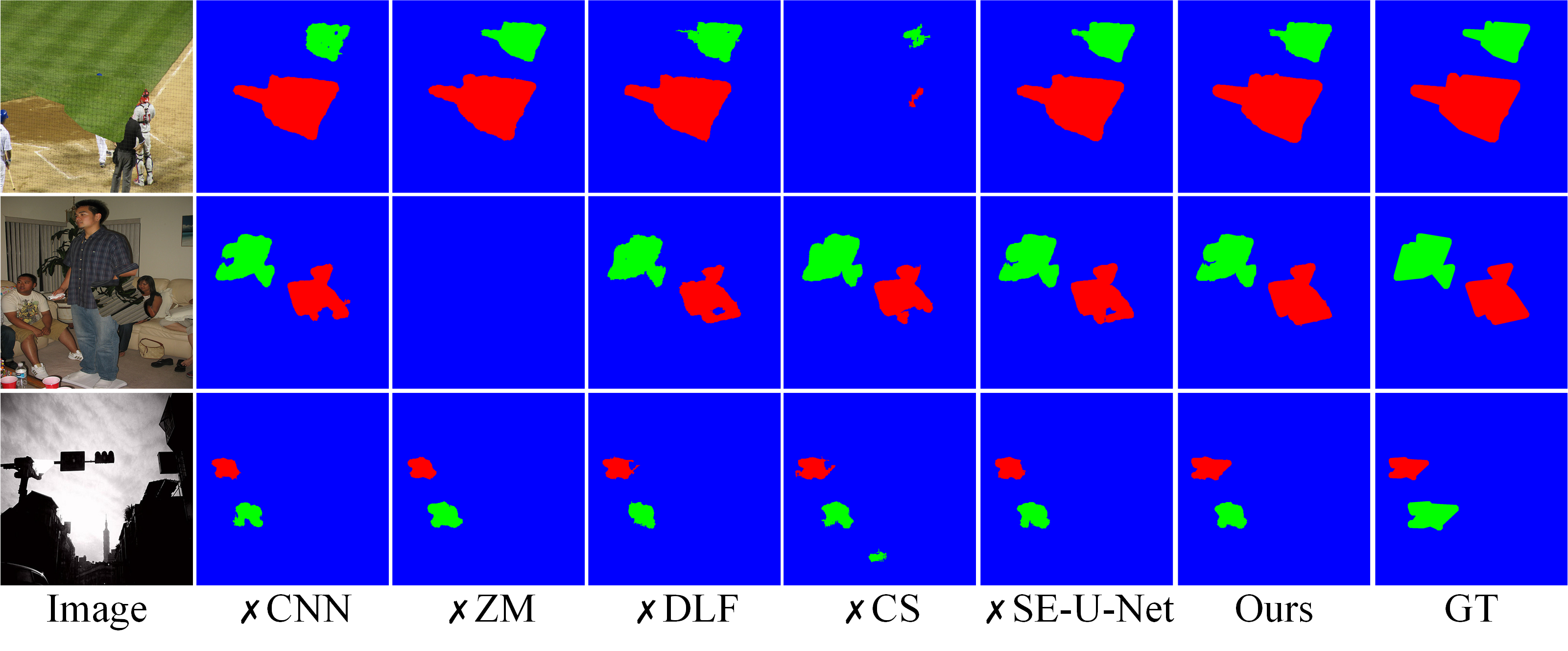}
	\centering
	\caption{Visual comparison of frameworks without specific modules.}
	\label{fig:AB_v}
\end{figure}

\subsubsection{Comparison with DisTool}
In this section, we compare our method with 
DisTool \cite{barni2020copy}, which focuses solely on source/target discrimination, assuming that a good binary localization mask has already been obtained. Additionally, DisTool imposes strict requirements on the input binary localization mask. For example, the mask must consist of two separate, non-contiguous regions (neither less nor more), and the ratio between these two regions cannot be too small. It's important to emphasize that in many cases, the binary localization masks obtained by existing models do not meet these criteria, rendering DisTool unable to process them. For further details, please refer to Section IV.D in the DisTool paper. 

Since DisTool relies on previously detected binary masks, we used the localization mask output from our DFM branch as the input for DisTool, comparing only the performance in source/target discrimination. Due to the limitations discussed above, DisTool was able to detect 3635 out of 4000 images in the synthetic dataset, 915 out of 1309 images in the CASIA dataset, and 100 out of 108 images in the CMH dataset. The comparison strategy aligns with the approach described in the DisTool paper, where the performance is computed \textit{only} on the images that DisTool can process. The results are shown in Table \ref{table:DisTool_1}. We can see that our proposed method outperforms DisTool in the source/target discrimination task, even when considering only those images that DisTool can process.
It is also important to note that our model can detect single-channel copy-move masks through an end-to-end process, a capability that DisTool lacks.

\subsection{Ablation Experiment}
In this subsection, we conduct ablation studies on our proposed model to assess the impact of each component on overall performance. Table \ref{table:ablation2} illustrates the step-by-step analyses to evaluate the effectiveness of each component in enhancing detection accuracy. In Table \ref{table:ablation2}, without CNN means removing the CNN branch of Feature Extraction I, without ZM means removing the Zernike Moment branch
of Feature Extraction I, without DLF means removing the Multi-scale DLF block, without CS-PM means removing the cross-scale matching strategy in the deep cross-scale PM, and without SE-U-Net means removing the branch of SE-U-Net.

Table \ref{table:ablation2} demonstrates that removing either CNN features or ZM features leads to a decrease in performance. This is because CNN features can adaptively learn highly discriminative task-specific features. Besides, we find that a pure CNN feature extraction from random initialization struggles to learn rotation-invariant features. Utilizing ZM features effectively addresses this issue and guides the model to quickly generate valid offsets in the early stages of training, enhancing training stability. Additionally, removing the DLF component results in about 7\% decrease in the F1 score for the source region and 10\% decrease for the target region.
Furthermore, it is also evident that cross-scale matching in deep PM significantly contributes to enhancing localization performance. Lastly, removing the SE-U-Net branch leads to a 5\% decrease in the F1 score for the target regions, aligning with our expectations and confirming the effective refinement of target area detection provided by SE-U-Net branch.

Fig. \ref{fig:AB_v} illustrates the visual results demonstrating the effectiveness of the modules discussed in Table \ref{table:ablation2} in improving the final detected masks. For instance, the cross-scale matching strategy (CS-PM) significantly enhances detection performance against large rescaling attacks. ZM features offer better rotational invariance capability, while the SE-U-Net module refines the target region in the final results.

\begin{figure}[t!]
	\centering
	\includegraphics[width=1\linewidth]{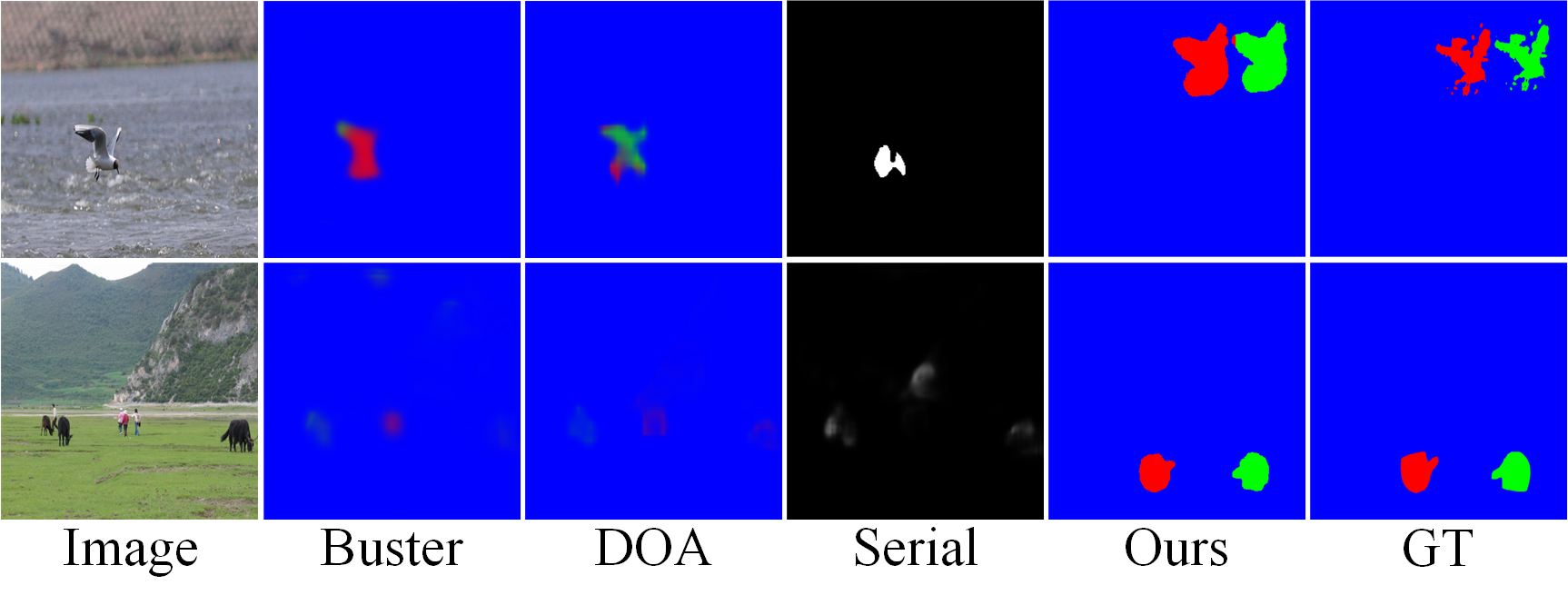}
	\centering
    \caption{Examples where} copy-move manipulation occurs in the background.
    \label{fig:background_challeng}
\end{figure}

\begin{figure}[t!]
	\centering
	\includegraphics[width=1\linewidth]{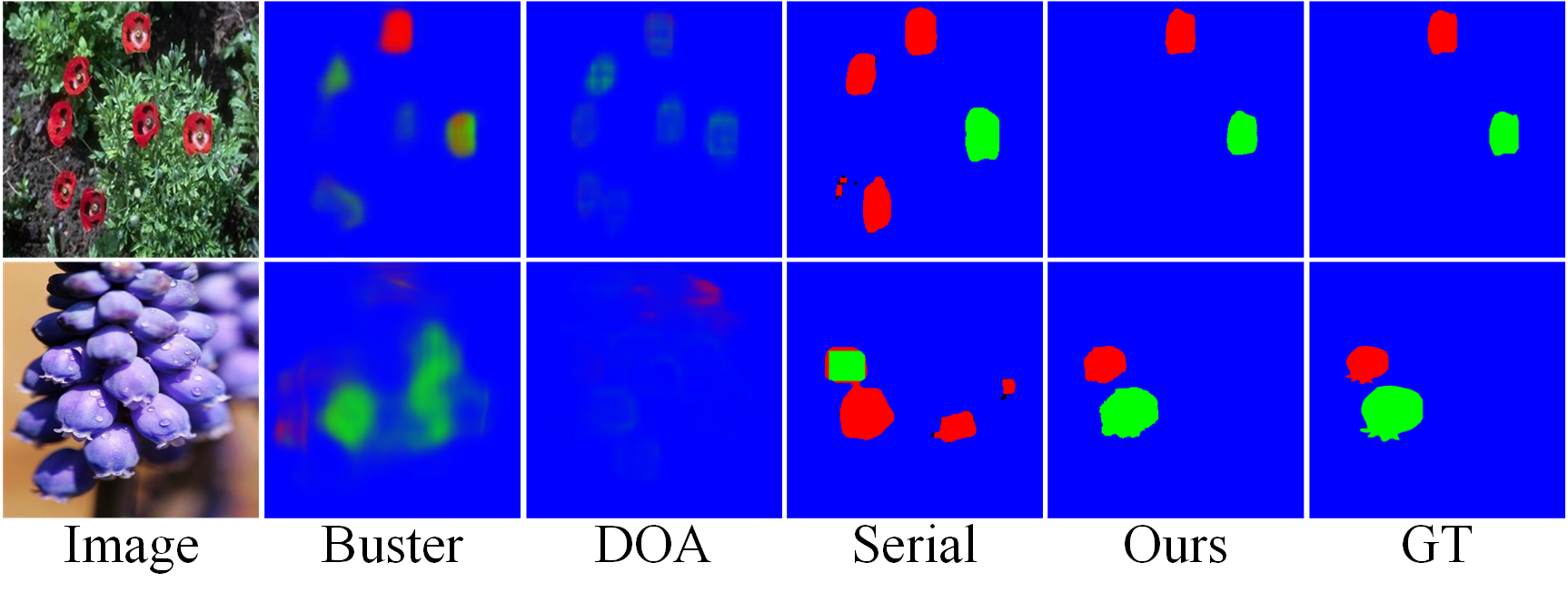}
	\centering
    \caption{Examples where} multiple similar objects exist in the background.
    \label{fig:similar_challeng}
\end{figure}

\begin{figure}[t!]
	\centering
	\includegraphics[width=1\linewidth]{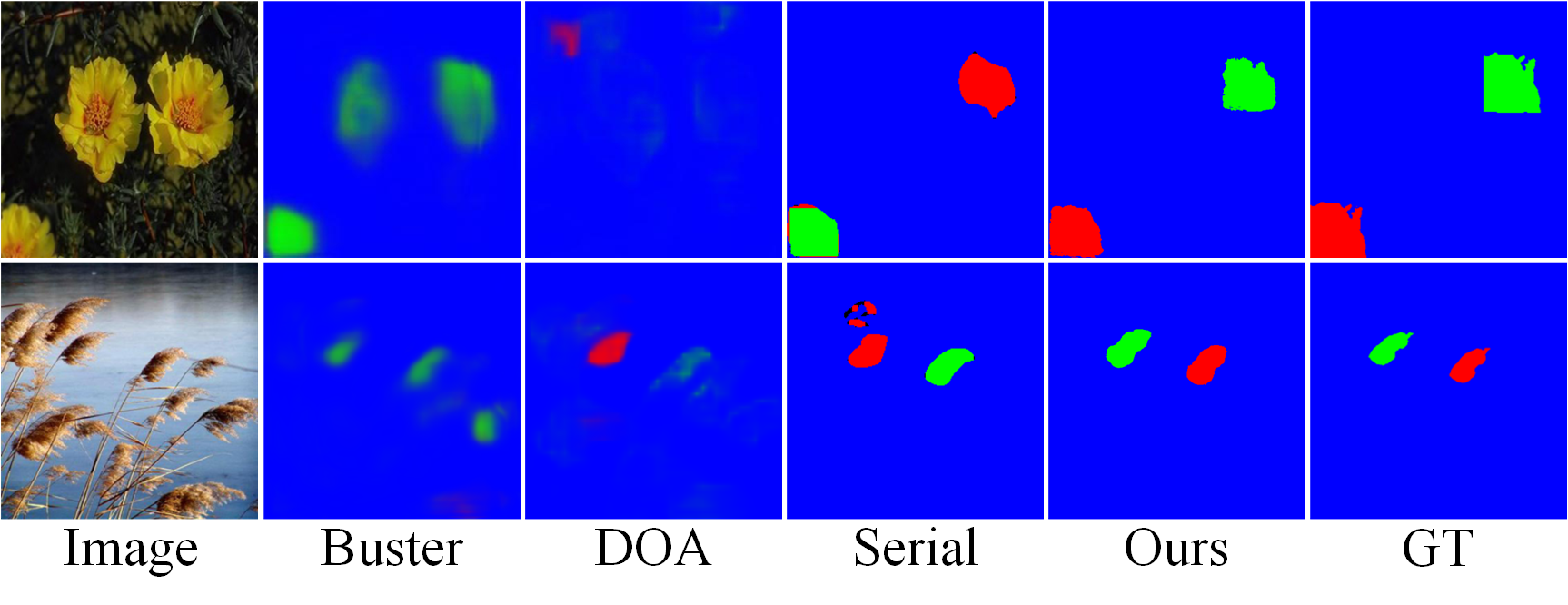}
	\centering
   \caption{Examples where target regions blend well with the background.}
    \label{fig:seamless_challeng}
\end{figure}

\subsection{Visual Comparisons on Some Challenging Cases}
To further demonstrate the effectiveness of our method, we present visual comparisons with challenging cases in this subsection.

\subsubsection{Detect forgeries where} copy-move manipulation occurs in the background
Fig. \ref{fig:background_challeng} presents two examples illustrating copy-move manipulations occurring only in the background. In the first case, a bird is concealed by a water background, representing a common copy-move scenario encountered in practice. The second case involves the copy-move of a grassland background region. In comparison to other deep methods, our approach effectively identifies genuine copy-move regions in the background and successfully discriminates between the source and target regions. Notably, other methods incorrectly respond to the region containing the bird in the first image. Similarly, in the second image, they tend to detect regions containing people and a cow. Our experimental results demonstrate that previous deep methods are likely tend to overfit objects, preventing them from accurately detecting the real copy-move regions.

\subsubsection{Detect forgeries where multiple similar objects exist in the background.}
Fig. \ref{fig:similar_challeng} illustrates another challenging scenario where the image contains multiple similar genuine objects. It is evident that previous deep models often respond to similar flowers and fruits in the image, but struggle to accurately distinguish between the source and target regions in most cases. Specifically, in the second image where the forgery traces are subtle, the DOA method fails to locate the copy-move region, and the Serial Net erroneously identifies the target region as the source region. In contrast, our proposed method successfully detects the copy-move regions based on low-level semantic features, effectively avoiding confusion about background objects by leveraging richer texture information.

\subsubsection{Source/target separation with seamless forging}
 Fig. \ref{fig:seamless_challeng} provides two examples to showcase our method's efficacy in accurately separating source and target regions even when the target regions seamlessly blend into the background. These scenarios involve very subtle forged traces. As observed, Busternet often labels all detected regions as the source region due to the faint forgery cues. Conversely, DOA and Serial Net yield opposite outcomes in distinguishing the source/target regions.
 However, it can be seen that our method achieves precise detection by leveraging piecewise rank learning. This approach enables our method to learn subtle forgery clues to separate the source and target pixels through a comparison of the matched features.

\section{Conclusion}\label{sec:Conculsion}
In this paper, we propose a novel end-to-end copy-move forgery detection framework with source/target discrimination. The proposed framework explicitly addresses the fundamental limitations of existing deep models, such as low generalizability to background forgery and poor discriminating ability for source-target separation. Specifically, we suggest a deep cross-scale PM algorithm and devise a pairwise ranking learning approach that effectively detects copy-move instances while uncovering subtle clues for distinguishing between the source and target regions. Experiments validate the remarkable performance of our method, even in challenging scenarios. 
Note that point-to-point matching forms the foundation for constructing affine matrices related to the source and target regions. A potential research direction is investigating differential affine estimation based on offsets and subsequently refining the identified regions. Additionally, extending our framework to video copy-move forgery detection could also be an intriguing topic for further exploration.

\bibliography{TD}
\bibliographystyle{IEEEtran}

\end{document}